\documentclass[lettersize,journal]{IEEEtran}
\usepackage{amsmath,amsfonts}
\usepackage{algorithmic}
\usepackage{array}
\usepackage[caption=false,font=normalsize,labelfont=sf,textfont=sf]{subfig}
\usepackage{textcomp}
\usepackage{stfloats}
\usepackage{url}
\usepackage{verbatim}
\usepackage{graphicx}
\usepackage{booktabs}
\usepackage{multirow}

\usepackage{color}
\usepackage{makecell}
\usepackage{caption}
\usepackage{adjustbox}
\usepackage{xcolor}
\usepackage{enumitem}
\usepackage[square, numbers]{natbib}
\usepackage[switch]{lineno}
\usepackage{amssymb}
\usepackage[
            colorlinks,
            linkcolor=blue,
            anchorcolor=blue,
            citecolor=blue,
            urlcolor=blue
            ]{hyperref}
            
\definecolor{additive}{RGB}{84, 130, 52}
\definecolor{selective}{RGB}{45, 117, 182}
\definecolor{prompt}{RGB}{192, 144, 1}
\definecolor{reparameter}{RGB}{183, 96, 41}
\definecolor{integrated}{RGB}{112, 47, 160}

\newcommand\iyes{\raisebox{-0.2em}{\includegraphics[width=1em]{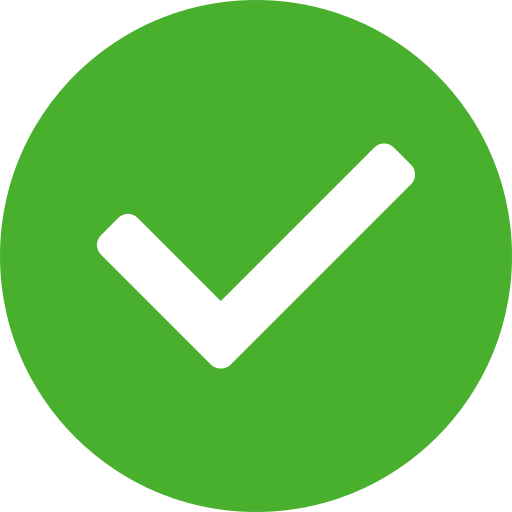}}}
\newcommand\ino{\raisebox{-0.2em}{\includegraphics[width=1em]{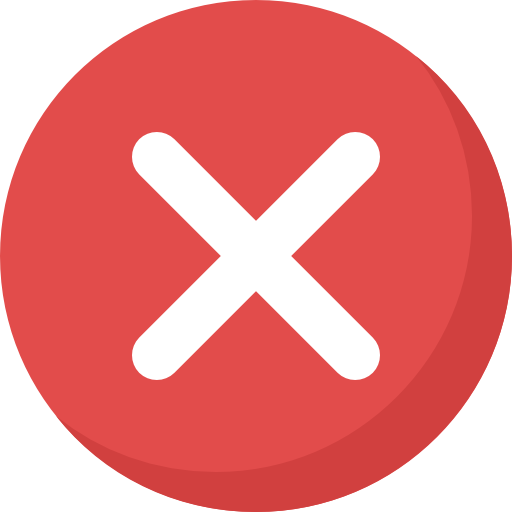}}}

\newcommand{\sssection}[1]{\vspace{0.2cm}\noindent\textit{#1}\vspace{0.2cm}}

\def\BibTeX{{\rm B\kern-.05em{\sc i\kern-.025em b}\kern-.08em
    T\kern-.1667em\lower.7ex\hbox{E}\kern-.125emX}}
\usepackage{balance}

\begin{document}
\title{Parameter-Efficient Fine-Tuning for \\ Foundation Models}

\author{
Dan Zhang$^*$, Tao Feng$^*$, Lilong Xue$^*$, Yuandong Wang, Yuxiao Dong, Jie Tang

The Knowledge Engineering Group (KEG), Tsinghua University

zd21@mails.tsinghua.edu.cn

\url{https://Awesome-PEFT-for-Foundation-Models.github.io}

\thanks{$^*$ Equal contributions.

Manuscript created October, 2020; This work was developed by the IEEE Publication Technology Department. This work is distributed under the \LaTeX \ Project Public License (LPPL) ( http://www.latex-project.org/ ) version 1.3. A copy of the LPPL, version 1.3, is included in the base \LaTeX \ documentation of all distributions of \LaTeX \ released 2003/12/01 or later. The opinions expressed here are entirely that of the author. No warranty is expressed or implied. User assumes all risk.}}

\markboth{Journal of \LaTeX\ Class Files,~Vol.~18, No.~9, September~2020}%
{How to Use the IEEEtran \LaTeX \ Templates}

\maketitle

\begin{abstract}
This survey delves into the realm of Parameter-Efficient Fine-Tuning (PEFT) within the context of Foundation Models (FMs). 
PEFT, a cost-effective fine-tuning technique, minimizes parameters and computational complexity while striving for optimal downstream task performance. FMs, like ChatGPT, DALL-E, and LLaVA specialize in language understanding, generative tasks, and multimodal tasks, trained on diverse datasets spanning text, images, and videos. 
The diversity of FMs guides various adaptation strategies for PEFT. Therefore, this survey aims to provide a comprehensive overview of PEFT techniques applied to diverse FMs and address critical gaps in understanding the techniques, trends, and applications. 
We start by providing a detailed development of FMs and PEFT. Subsequently, we systematically review the key categories and core mechanisms of PEFT across diverse FMs to offer a comprehensive understanding of trends. We also explore the most recent applications across various FMs to demonstrate the versatility of PEFT, shedding light on the integration of systematic PEFT methods with a range of FMs.
Furthermore, we identify potential research and development directions for improving PEFTs in the future. This survey provides a valuable resource for both newcomers and experts seeking to understand and use the power of PEFT across FMs.
All reviewed papers are listed at~\url{https://github.com/THUDM/Awesome-Parameter-Efficient-Fine-Tuning-for-Foundation-Models}.
\end{abstract}


\section{Introduction}
\label{sec: intro}

\begin{figure*}[t!]
    \centering
    \includegraphics[width=1.00\linewidth]{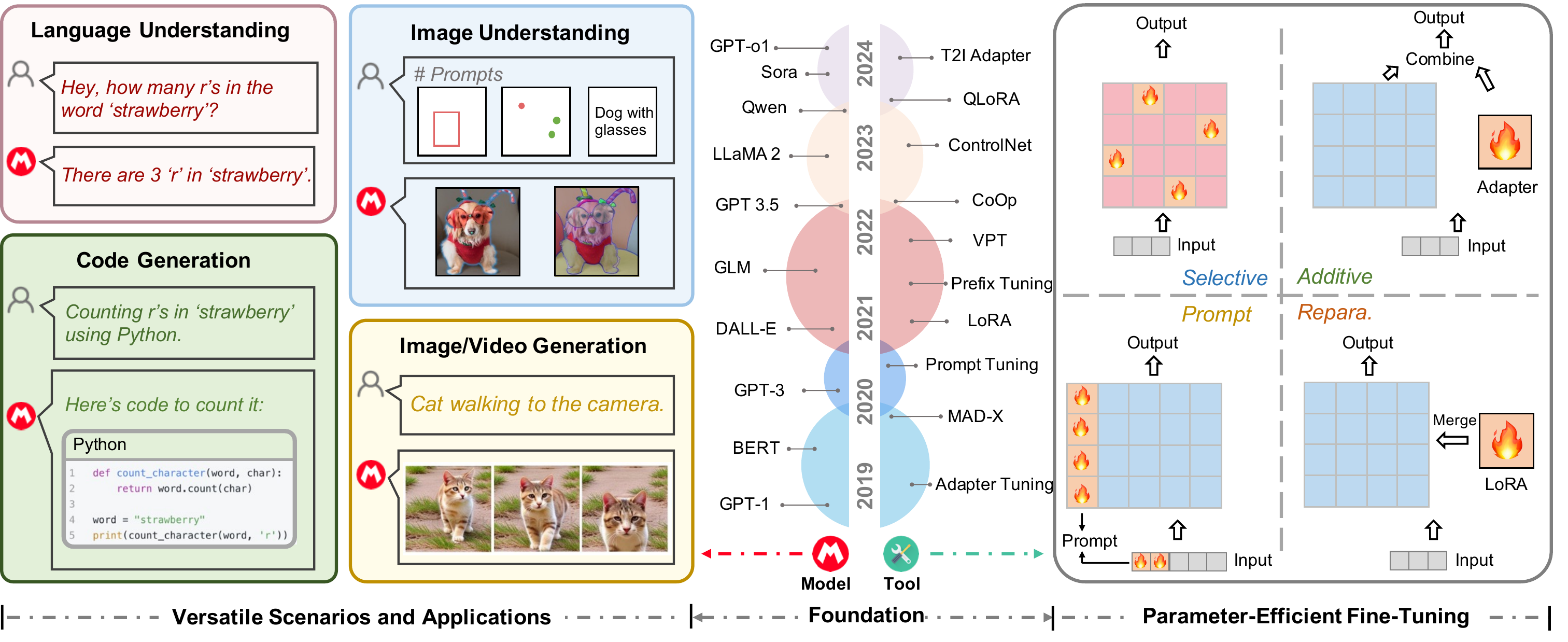}
    \caption{Left: Versatile scenarios and applications in the era of FMs. Right: A detailed illustration of four common PEFT methods (\textcolor{selective}{Selective}, \textcolor{additive}{Additive}, \textcolor{prompt}{Prompt}, and \textcolor{reparameter}{Reparameterization} PEFT).}
    \label{fig: framework}
\end{figure*}

Foundation Models (FMs) are pre-trained on large-scale datasets~\cite{Radford2018ImprovingLU, Radford2019LanguageMA, Brown2020LanguageMA, OpenAI2023GPT4TR, zheng2023judging, Touvron2023LLaMAOA} (often covering various types, such as text, images, and videos, etc.) to cater to multiple tasks like language understanding~\cite{liu2019text, he2022z, liu2012emoticon, zhang2022incorporating, zhang2023prompting, conneau2019cross, shao2023prompting, yu2024self, zhang2024sciglm, zhang2024rest, xue2024autore}, code generation~\cite{zheng2023codegeex, xia2024scenegenagent}, image or video understanding~\cite{ding2021cogview}, visual content generation~\cite{xu2024imagereward, hong2022cogvideo, yang2024cogvideox}, as depicted in Fig.~\ref{fig: framework} (left). Presently, various FMs dominate distinct domains, for instance, language-focused tasks are supported by ChatGPT~\cite{OpenAI2023GPT4TR}, ChatGLM~\cite{zeng2022glm, glm2024chatglm}, and Qwen~\cite{bai2023qwen}, while vision-language tasks are tackled by ChatGPT-4V~\cite{yang2023dawn}. 
DALL-E~\cite{ramesh2021zero}, Sora~\cite{liu2024sora}, and Veo2\footnote{https://deepmind.google/technologies/veo/veo-2/} specialize in generative tasks, and LLaVA~\cite{LLaVA}, and NExT-GPT~\cite{wu2023next} excel at multimodal ones, as depicted in Fig.~\ref{fig: framework} (middle). In real-world applications, fine-tuning these FMs on unseen downstream datasets is usually required to achieve task-specific performance.

\begin{figure}[t!]
    \centering
    \includegraphics[width=1.00\linewidth]{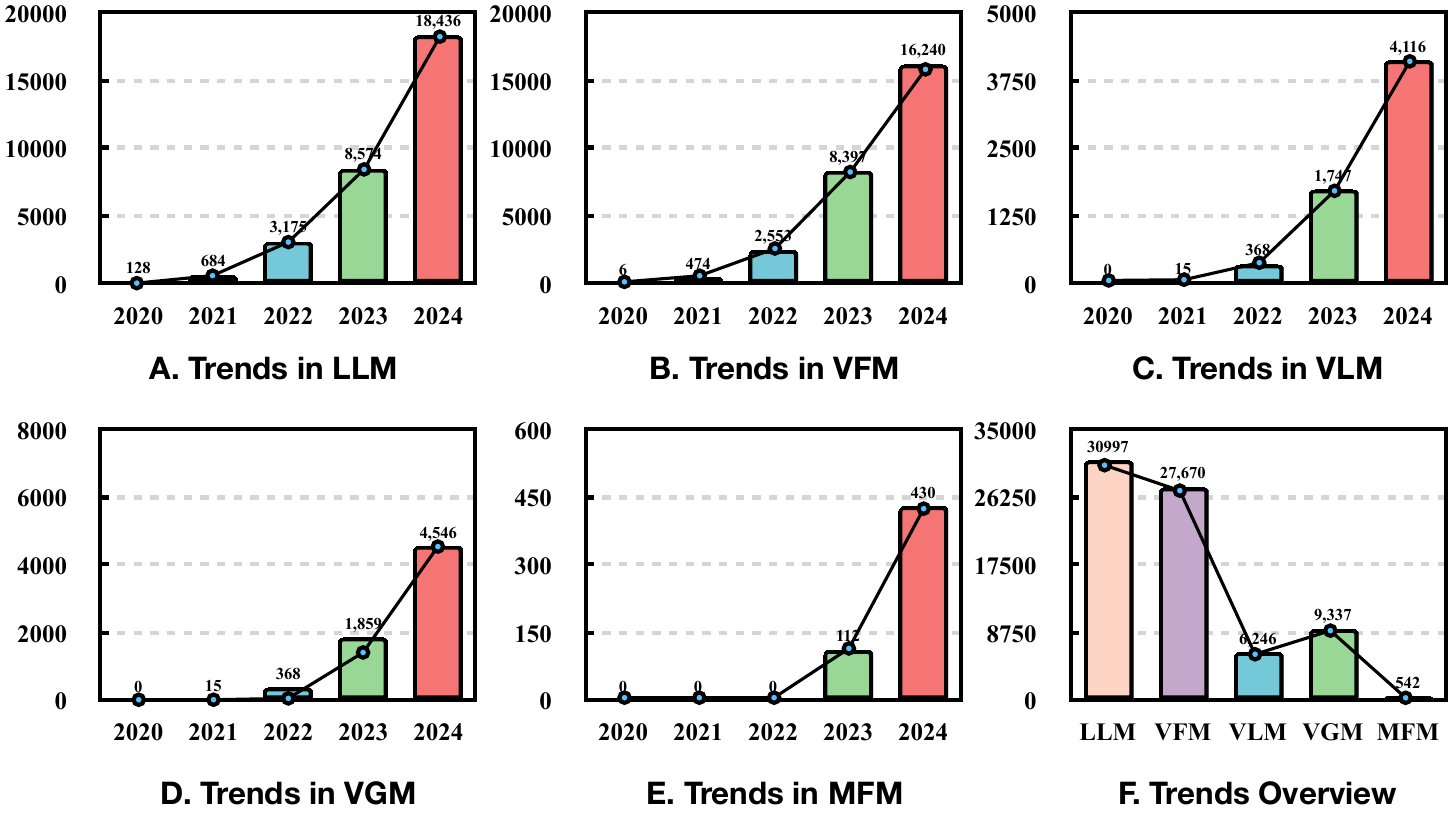}
    \caption{An overview of trends in PEFT methods in various FMs (LLM, VFM, VLM, MFM, and VGM). The number of citations from Semantic Scholar serves as a trend indicator.}
    \label{fig: trends}
\end{figure}

Parameter-Efficient Fine-Tuning (PEFT) technology~\cite{xin2024parametervision, han2024parameterlarge, zhou2024empirical, wang202410parameter}, a highly active topic, demonstrates notable cost-effectiveness during the fine-tuning process, as depicted in Fig.~\ref{fig: trends} and Fig.~\ref{fig: framework} (right). This technique minimizes the trainable parameters and computational overhead while aspiring to near fully fine-tuned performance on downstream tasks. 
Taking GPT-3~\cite{Brown2020LanguageMA} as an example, full fine-tuning involves all 175B parameters, whereas LoRA~\cite{Hu2021LoRALA} requires training only 4.7M or 37.7M, saving over 99.97\% of parameters, and the result is a 0.1\% to 0.5\% improvement compared to full fine-tuning.
Such attributes brought significant practical value to the community and real-world applications\footnote{https://github.com/huggingface/peft?tab=readme-ov-file\#high-performance-on-consumer-hardware}. Nonetheless‌, the diversity of FMs steered the various adaptation strategies for PEFT. For example, in the prompt tuning approach, the design of trainable prompt modules often varies depending on the type of FMs (e.g., text prompts~\cite{Liu2021PTuningVP} for large language models (LLMs), and visual prompts~\cite{Jia2022VisualPT} for vision language models (VLMs)). Similarly, LoRA~\cite{Hu2021LoRALA} is integrated into different components of FMs depending on their architecture (e.g., transformer blocks~\cite{vaswani2017attention} for LLMs or denoising U-Net~\cite{gurrola2021residual} for vision content generation models (VGMs)). Consequently, conducting a comprehensive survey of how PEFT techniques are adapted across diverse FMs is crucial for advancing this field. This understanding will pave the way for more systematic and effective applications of PEFT across a wide range of tasks and domains.

As highlighted above, FMs are iterating at an unprecedented pace concerning structure, method, and applications. This rapid evolution fueled the PEFT community to become equally dynamic. Hence, keeping track on the technological trend of PEFT within FMs is imperative. As shown in Fig.~\ref{fig: trends}, we count the total number of citations of PEFT methods across various FMs over the past five years as a trend indicator and have three key trends: 

\textit{Trend i: The field of PEFT is experiencing remarkable growth, covering a diverse range of tasks and FMs, including language, vision, and multimodal domains.}

\textit{Trend ii: LLMs and vision foundation models (VFMs) dominate the current research landscape, showing rapid and substantial increases in activity, while VLMs and vision content generation models (VGMs) are gaining traction as secondary areas of focus.}

\textit{Trend iii: In contrast, multimodal foundation models (MFMs) remain relatively underexplored, suggesting significant opportunities for future research and innovation in this area.}

In this survey, we aim to explore the potential of integrating PEFT with various FMs to enhance scalability. 
Furthermore, given the mutual dynamism of these two communities, several overview surveys recently emerged, as shown in Table~\ref{tab: comparison}. Like, Xin et al.~\cite{xin2024parametervision} systematically review visual PEFT (covering common datasets and applications) while identifying future directions. Zhou et al.~\cite{zhou2024empirical} expand the scope to multimodal LLMs and present empirical studies of several mainstream PEFT methods. Their findings highlight the superior performance of adapter tuning and the positive role of connector layers in fine-tuning MFMs. Wang et al.~\cite{wang202410parameter} focused on the core ideas and principles of various PEFT algorithms, providing a quick theoretical guide. Notably, Han et al.~\cite{han2024parameterlarge} offered detailed insights into PEFT for LLMs from an algorithmic standpoint and proposed recommendations for system design in real-world scenarios. These valuable surveys offer focused insights into some lines of PEFT. However, these insights are scattered across different studies for generalized FMs. Second, this field lacks close attention to the developmental lines of PEFT across various FMs and a more intuitive and unified illustration. Thus, a well-structured and comprehensive survey has become increasingly necessary.

Therefore, we first review the trends in FM development and the categorization of PEFT (Section~\ref{sec: background}). Subsequently, we delve into the design of PEFT across five model structures (Section~\ref{sec: method}), including \textcolor{selective}{Selective} PEFT, \textcolor{additive}{Additive} PEFT, \textcolor{prompt}{Prompt} PEFT, \textcolor{reparameter}{Reparameterization} PEFT, and \textcolor{integrated}{Hybrid} PEFT, with corresponding feature summary in Table~\ref{tab: methods}. We also explore the applications of PEFT in different downstream tasks and their corresponding scenarios (Section~\ref{sec: application_llm} for LLMs, Section~\ref{sec: application_vis_found} for VFMs, and Section~\ref{sec: application_mm} for MFMs). Finally, we provide observations on current research trends and potential future research directions (Section~\ref{sec: discussion}) to aid in the community development of PEFT across various domains. Through this survey, we provide a deeper understanding of the integration between a wide range of FMs and systematic PEFT methods.

\begin{table*}[t!]
    \centering
    \renewcommand{\arraystretch}{1.0}

    \begin{adjustbox}{max width=\linewidth}
    \begin{tabular}{c|c|c|c|c|c|c|c|c}
    \specialrule{.16em}{0pt} {.65ex}
        \multirow{2}{*}{\textbf{Survey}} & \multirow{2}{*}{\textbf{Venue}} & \multicolumn{4}{c|}{\begin{tabular}[c]{@{}c@{}}\textbf{FMs} {(Fig. \ref{fig: framework}, \ref{fig: adapter}, \ref{fig: Prompt}, \ref{fig: diff_lora}, \ref{fig: figure_SD})}\end{tabular}} & \multirow{2}{*}{\begin{tabular}[c]{@{}c@{}}\textbf{Trend}\\ {(Fig. \ref{fig: trends})}\end{tabular}} &  \multirow{2}{*}{\begin{tabular}[c]{@{}c@{}}\textbf{Stats.}\\ {(Tab. \ref{tab: methods})}\end{tabular}} & \multirow{2}{*}{\textbf{Pros\&Cons}} \\
        \cmidrule{3-6}
         & & LLM & VFM & VLM/MFM & VGM & & \\
         \specialrule{.05em}{.4ex}{.65ex}
         Xin et al.~\cite{xin2024parametervision} & arXiv, 2024    &\ino  &\iyes &\ino  &\iyes  &\ino &\ino &\ino \\ 
         Han et al.~\cite{han2024parameterlarge} & arXiv, 2024    &\iyes  &\iyes  &\iyes  &\iyes  &\ino &\ino &\ino \\ 
         Zhou et al.~\cite{zhou2024empirical} & arXiv, 2024     &\ino  &\ino  &\iyes  &\ino  &\ino &\ino &\ino \\
         Wang et al.~\cite{wang202410parameter} & arXiv, 2024      &\iyes  &\iyes  &\iyes  &\iyes  &\ino &\ino &\ino \\
         \specialrule{.10em}{.4ex}{.65ex}
         Ours & arXiv, 2025  &\iyes  &\iyes  &\iyes  &\iyes  &\iyes  &\iyes &\iyes \\
        \bottomrule
    \end{tabular}
    \end{adjustbox}
    \caption{Summary of representative surveys on PEFT.
    \textbf{Trend, Stats., and Pros\&Cons} represents whether to provide trend analysis, statistics on the number of trainable parameters, and Pros and Cons analysis.}
    \label{tab: comparison}
\end{table*}
\section{Background}
\label{sec: background}
\subsection{Overview of Foundation Models}

FMs are primarily pre-trained on large-scale datasets and can be fine-tuned to adapt to various downstream tasks. Following the differences in their input modalities and functions, we roughly categorized them into five groups.

\textbf{Large Language Model (LLM)} is designed to understand, generate, and manipulate text. These models are trained on vast amounts of text corpora and can perform a wide range of language-related tasks, such as translation, summarization, text generation, and question-answering. Like BERT~\cite{devlin2018bert}, LLaMA~\cite{Touvron2023LLaMAOA}, GPT-4~\cite{OpenAI2023GPT4TR}, and ChatGLM~\cite{glm2024chatglm}.

\textbf{Vision Foundation Model (VFM)} focuses on understanding and generating insights from visual data, such as images. They can handle tasks such as image classification, object detection, segmentation, and more. The models are pre-trained on large-scale image datasets, allowing them to generalize well to a variety of vision-related tasks. Like Grounding DINO~\cite{liu2023grounding} and SAM~\cite{kirillov2023segany}.

\textbf{Vision Language Model (VLM)} integrates both visual and textual modalities, enabling tasks that require understanding the relationship between images and language. They are used in applications such as grounding, image captioning, and visual question answering. Like CLIP~\cite{radford2021learning}, BLIP~\cite{li2022blip}, GPT-4V~\cite{yang2023dawn}, and GLM-4V~\cite{hong2024cogvlm2}.

\textbf{Visual Content Generation Model (VGM)} focuses on generating high-quality visual content, such as images, videos, or 3D models, from various inputs (text, sketches, or other visual prompts). They are used in art generation, video synthesis, and even in creating synthetic training data for other AI models. Like Stable Diffusion~\cite{rombach2021highresolution}, DALL-E~\cite{ramesh2021zero}, Zero-1-to-3~\cite{liu2023zero1to3}, and CogVideo-X~\cite{yang2024cogvideox}.

\textbf{Multi-Modal Foundation Model (MFM)} extends the capabilities of LLMs to handle multiple modalities, such as text, images, and sometimes audio. These models can simultaneously process and generate text, images, and audio, etc., enabling richer interactions in multi-modal tasks. Like LLaVA-1.5~\cite{LLaVA}, Gemini 1.5 Pro~\cite{reid2024gemini}, CoDi~\cite{tang2024any, tang2024codi}, SEED-X~\cite{ge2024seed}, and NExT-GPT~\cite{wu2023next}.

\subsection{Development of Parameter-Efficient Fine-Tuning}
PEFT has emerged as a significant approach for fine-tuning foundation models (such as BERT and GPT-3), aiming to reduce the number of parameters that need to be updated during the tuning process, thereby lowering computational and storage costs. Below is a summary description of the key PEFT developments and associated methods:

\textbf{\textcolor{selective}{Selective} PEFT.} This category focuses on fine-tuning only a subset of the model's parameters instead of all of them. The fundamental assumption here is that certain parameters are particularly important for specific tasks in large pre-trained models, and adjusting these key parameters can yield satisfactory results.
Early methods like layer-wise freezing~\cite{Lee2019WhatWE} gradually thaw the layers of the model during the fine-tuning process. 
More partial strategies~\cite{Liu2021AutoFreezeAF, Xu2021RaiseAC} have also emerged, identifying which layers should be thawed and adjusted through either empirical methods or learning processes.

\textbf{\textcolor{additive}{Additive} PEFT.} Additive methods involve inserting small adapter networks, also known as bottleneck adapters~\cite{Houlsby2019ParameterEfficientTL}, between the layers of FMs, making minimal changes to achieve fine-tuning.
One of the earliest adapter methods inserts bottleneck layers between the model layers, updating these bottleneck parameters while keeping the original model largely unchanged. Adapters~\cite{Pfeiffer2020AdapterFusionNT, Sung2022LSTLS, jie2022convolutional} significantly reduce the number of parameters that need to be updated.

\textbf{\textcolor{prompt}{Prompt} PEFT.} This category involves learning soft commands~\cite{Li2021PrefixTuningOC, Liu2021GPTUT}, sequences of embedding vectors~\cite{Gu2021PPTPP}, that guide the model to perform the task effectively.

\textbf{\textcolor{reparameter}{Reparameterization} PEFT.} These methods~\cite{Hu2021LoRALA, Zi2023DeltaLoRAFH, gu2024mix} propose re-representing or decomposing existing model parameters so that only part of them need to be adjusted during fine-tuning, thereby preserving the majority of the unchanged parameters.

\textbf{\textcolor{integrated}{Hybrid} PEFT.} These methods~\cite{Mao2021UniPELTAU, Chen2023ParameterEfficientFD, zhang2022neural} combine multiple PEFT strategies to achieve optimal results, incorporating techniques such as adapters, prompts, and parameterizations. 
Recent approaches focus on finding the best configuration of these strategies for different tasks and scenarios.

In summary, the evolution of PEFT is characterized by diversification and integration, as shown in Fig.~\ref{fig: trends} and Fig.~\ref{fig: framework}. 
As model sizes continue to increase and multitask learning demands grow, PEFT methods constantly evolve, providing innovative solutions for more efficient and resource-saving model fine-tuning.

\begin{table*}[!t]
\small
\centering
\renewcommand{\arraystretch}{1.0}
\begin{adjustbox}{max width=0.95\linewidth}
\begin{tabular}{c|c|c|c|c|c|c|c|c}

\specialrule{.16em}{0pt} {.65ex}
\textbf{Approach} & \textbf{Venue} & \textbf{FMs} & \textbf{Category}  & \textbf{SC} & \textbf{Position} & \textbf{IE} & \textbf{Add. Params.} & \begin{tabular}[c]{@{}c@{}}\textbf{Trainable}\\\textbf{Parameters}\end{tabular}   \\ 

\specialrule{.05em}{.4ex}{.65ex}

Freeze Layers~\citep{Lee2019WhatWE} & arXiv, 2019 &LLM& \textcolor{selective}{S} & \ino  & Layers & \iyes & \ino & 1/4 final layers \\
 
Masking~\citep{Zhao2020MaskingAA}& arXiv, 2020 &LLM& \textcolor{selective}{S} & \ino  & - & \iyes &   \ino & 3\%-10\%\\
 
Diff Pruning~\citep{Guo2020ParameterEfficientTL}& ACL, 2020 &LLM&\textcolor{selective}{S} & \ino  & - & \iyes & \ino   & 0.5\% \\
  
CHILD-TUNING~\citep{Xu2021RaiseAC} & EMNLP, 2021 &LLM& \textcolor{selective}{S} & \ino  & - & \iyes &   \ino  & 0.1\% - 0.4\%  \\
  
FISH~\citep{Sung2021TrainingNN} & NeurIPS, 2021 &LLM& \textcolor{selective}{S} & \ino  & - & \iyes &   \ino & 0.5\% \\
 
BitFit~\citep{BenZaken2021BitFitSP} & ACL, 2022 & LLM & \textcolor{selective}{S} & \ino  & Attention & \iyes &  \ino & 0.01\% - 0.09\%  \\

PASTA~\citep{Yang2022ParameterEfficientTW} & arXiv, 2022 & LLM & \textcolor{selective}{S} & \ino  & Input & \iyes & \ino & 0.029\%  \\
 
LT-SFT~\citep{Ansell2021ComposableSF} & ACL, 2022 & LLM& \textcolor{selective}{S} & \ino  & - & \iyes &   \ino &  \begin{tabular}[c]{@{}c@{}}0.16\% - 8\%\end{tabular}
  \\ 
  
FC-CLIP~\cite{yu2024convolutions} & NeurIPS, 2023 & VFM & \textcolor{selective}{S} & \ino  & Classifier & \iyes &  \ino & 1.8\% - 8\%\\
  
Tune-A-Video~\cite{wu2023tune} & ICCV, 2023 & VGM & \textcolor{selective}{S} & \ino  & U-Net & \iyes & \ino &- \\

LayerNorm Tuning~\cite{zhao2023tuning} & ICLR, 2024 & MFM & \textcolor{selective}{S} & \ino  & LayerNorm & \iyes &  \ino & 2.5\% - 3.78\% \\

\specialrule{.10em}{.4ex}{.65ex}

Bottleneck Adapter~\citep{Houlsby2019ParameterEfficientTL} & arXiv, 2019 &LLM & \textcolor{additive}{A} & \iyes  & FFN & \ino &    \iyes   &  3.6\%  \\   
 
MAD-X ~\citep{Pfeiffer2020MADXAA} & EMNLP, 2020&LLM& \textcolor{additive}{A} & \iyes  & FFN & \ino &   \iyes  & 3.05\%  \\ 

AdaMix~\citep{Wang2022AdaMixMF} & EMNLP, 2022 &LLM& \textcolor{additive}{A} & \iyes  & FFN & \ino &   \iyes & 0.1\% - 0.2\% \\ 

AdapterBias~\citep{Fu2022AdapterBiasPT} & NAACL, 2022 &LLM& \textcolor{additive}{A} & \iyes  & FFN & \ino &   \iyes 
&  0.04\% - 0.05\%\\ 

LST~\citep{Sung2022LSTLS} & NeurIPS, 2022 &LLM/VLM& \textcolor{additive}{A} & \iyes  & Side Network & \ino &   \iyes & 0.08\% - 7.46\% \\ 

Convpass~\cite{jie2022convolutional} & arXiv, 2022 & VFM & \textcolor{additive}{A} & \iyes  & MSA\&FFN & \ino &  \iyes & 0.19\% - 0.38\% \\ 

AdaptFormer~\cite{chen2022adaptformer} & NeurIPS, 2022 & VFM & \textcolor{additive}{A} & \iyes  & MLP & \ino &  \iyes & 0.18\%\\ 

ViT-Adapter~\cite{chen2022vision} & ICLR, 2023& VFM & \textcolor{additive}{A} & \iyes  & ViT & \ino & \iyes& -\\ 

SAN~\cite{xu2023side} & CVPR, 2023 & VFM & \textcolor{additive}{A} & \iyes  & ViT & \ino &  \iyes & 4.47\% \\ 

FacT~\cite{jie2023fact} & AAAI, 2023 & VFM & \textcolor{additive}{A} & \iyes  & MSA\&FFN & \ino &  \iyes & 0.08\% \\
CSN (DTL)~\cite{fu2024dtl} & AAAI, 2024 & VFM & \textcolor{additive}{A} & \iyes  & ViT & \ino &  \iyes & 0.05\% - 0.06\% \\

IP-Adapter~\cite{ye2023ip} & arXiv, 2023 & VGM & \textcolor{additive}{A} & \iyes  & U-Net & \ino &  \iyes & 7.23\% \\ 
ControlNet~\cite{zhang2023adding} & ICCV, 2023 & VGM & \textcolor{additive}{A} & \iyes  & U-Net & \ino &  \iyes & 22.8\% - 24.7\% \\ 
T2I-Adapter~\cite{mou2024t2i} & AAAI, 2024 & VGM & \textcolor{additive}{A} & \iyes  & U-Net & \ino & \iyes & -\\ 

I2V-Adapter~\cite{guo2024i2v} & SIGGRAPH, 2024 & VGM & \textcolor{additive}{A} & \iyes  & Attention & \ino &  \iyes& 1\% \\ 

ControlNeXt~\cite{peng2024controlnext} & arXiv, 2024 & VGM & \textcolor{additive}{A} & \iyes  & U-Net & \ino &  \iyes & 3.49\% - 4.06\% \\ 

LLaMA Adapter V2~\cite{gao2023llama} & ICLR, 2024 &  MFM & \textcolor{additive}{A} & \iyes  & - & \ino &  \iyes& 0.0006\%\\ 

Tip-Adapter~\cite{zhang2021tip} & ECCV, 2022 & VLM & \textcolor{additive}{A} & \iyes & CLIP & \ino & \iyes & - \\

CLIP-Adapter~\cite{gao2024clip} & IJCV, 2024 & VLM & \textcolor{additive}{A} & \iyes  & CLIP & \ino &  \iyes& 2 linear layers \\ 
\specialrule{.10em}{.4ex}{.65ex}
Prompt-Tuning~\cite{Lester2021ThePO} & EMNLP, 2021 & LLM& \textcolor{prompt}{P} & \ino & Input & \ino & \iyes & 20,480 params (5 tokens)  \\ 

Null Prompts~\citep{LoganIV2021CuttingDO} & arXiv, 2021& LLM & \textcolor{prompt}{P} & \ino & Input & \ino & \iyes 
&  0.1\%  \\

Prefix-Tuning~\cite{Li2021PrefixTuningOC} & ACL, 2021 & LLM & \textcolor{prompt}{P} & \iyes & Attention & \ino & \iyes &  0.1\%  \\ 

PPT~\citep{Gu2021PPTPP} & ACL, 2021 & LLM& \textcolor{prompt}{P} & \ino & Input & \ino & \iyes & 0.004\%  \\ 

SPoT~\citep{Vu2021SPoTBF} & ACL, 2021 & LLM&  \textcolor{prompt}{P} & \ino & Input & \ino & \iyes & 0.003\% \\ 

VP~\cite{bahng2022exploring} & arXiv, 2022 &VFM/VLM & \textcolor{prompt}{P} & \ino & Input & \ino & \iyes& 0.05\% - 4.4\% \\
VPT~\cite{jia2022visual} & ECCV, 2022 &VFM & \textcolor{prompt}{P} & \iyes & Input & \ino & \iyes& 0.04\% - 4.9\% \\
DAM-VP~\cite{huang2023diversity} & CVPR, 2023 &VFM & \textcolor{prompt}{P} & \ino & Input & \ino & \iyes& 6.3\% \\
ILM-VP~\cite{chen2023understanding} & CVPR, 2023 & VFM & \textcolor{prompt}{P} & \ino & Input & \ino & \iyes& 0.06\% - 0.43\%\\
EVP~\cite{wu2022unleashing} & TMLR, 2024 & VFM & \textcolor{prompt}{P} & \ino & Input & \ino & \iyes& 0.04\%\\
LION~\cite{wang2024lion} & AAAI, 2024 & VFM & \textcolor{prompt}{P} & \ino & Input\&Output & \ino & \iyes& 0.14\% - 0.41\% \\

Textual Inversion~\cite{gal2022image} & ICLR, 2023 & VGM & \textcolor{prompt}{P} & \ino & Input & \ino & \iyes& - \\

CoOp~\cite{zhou2022learning} & IJCV, 2022 & VLM & \textcolor{prompt}{P} & \ino & Input & \ino & \iyes & - \\
OVSeg~\cite{liang2023open} & CVPR, 2023 & VLM & \textcolor{prompt}{P} & \ino & Input & \ino & \iyes &- \\
Q-Former~\cite{li2023blip} & ICML, 2023 & MFM/VLM & \textcolor{prompt}{P} & \ino & - & \ino & \iyes & 0.89\% - 3.35\% \\

\specialrule{.10em}{.4ex}{.65ex}

LoRA~\citep{Hu2021LoRALA} & ICLR, 2021 & LLM& \textcolor{reparameter}{R} & \ino & Attention & \iyes & \iyes & 0.02\% - 0.31\% \\ 

MPO~\citep{Liu2021EnablingLF} & ACL, 2021 & LLM & \textcolor{reparameter}{R} & \ino & Attention & \iyes & \iyes & 9\%  \\ 

LoRA-FA~\citep{zhang2023lora} & arXiv, 2023 & LLM & \textcolor{reparameter}{R} & \ino & Attention & \iyes & \iyes & 1.5\% \\ 

IncreLoRA~\citep{zhang2023increlora} & arXiv, 2023 & LLM & \textcolor{reparameter}{R} & \ino & Attention & \iyes & \iyes &  0.01\% - 0.5\%\\ 

Delta-LoRA~\citep{Zi2023DeltaLoRAFH} & arXiv, 2023 & LLM & \textcolor{reparameter}{R} & \ino & Attention & \iyes & \iyes & 0.1\% \\ 

KronA~\citep{Edalati2022KronAPE} & NeurIPSW, 2024 & LLM&  \textcolor{reparameter}{R} & \ino & Attention & \iyes & \iyes & 0.07\%  \\

LoRand~\cite{yin20231} & CVPR, 2023 & VFM & \textcolor{reparameter}{R} & \ino & MSA\&FFN & \iyes & \iyes & 1.84\% - 2.76\%  \\

LyCORIS~\cite{yeh2023navigating} & ICLR, 2023 & VGM & \textcolor{reparameter}{R} & \ino & U-Net & \iyes & \iyes & - \\
DiffuseKronA~\cite{marjit2024diffusekrona} & arXiv, 2024 & VGM & \textcolor{reparameter}{R} & \ino & U-Net & \iyes & \iyes & 0.05\% - 65\% \\
T-LoRA (Customize-A-Video)~\cite{ren2024customize} & ECCV, 2024 & VGM & \textcolor{reparameter}{R} & \ino & 3D U-Net & \iyes & \iyes& -\\
ED-LoRA (Mix-of-Show)~\cite{gu2024mix} & NeurIPS, 2024 & VGM & \textcolor{reparameter}{R} & \ino & U-Net & \iyes & \iyes& -\\

LoRA-Sparse~\cite{song2024low} & CVPR, 2024 & MFM &\textcolor{reparameter}{R} & \ino & Attention & \iyes & \iyes & -\\

\specialrule{.10em}{.4ex}{.65ex}

COMPACTER~\citep{Davison2021CompacterEL} & arXiv, 2021 & LLM & \textcolor{integrated}{H} & \iyes & - & \ino & \iyes  & 0.047\%  \\

MAM~\citep{He2021TowardsAU} & ICLR, 2021 & LLM & \textcolor{integrated}{H} & \iyes & - & \ino & \iyes &0.5\% - 12.3\%   \\ 

UniPELT~\citep{Mao2021UniPELTAU} & ACL, 2022 & LLM &  \textcolor{integrated}{H} & \iyes & - & \ino & \iyes & 0.99\% - 1.26\%  \\ 

${S^4}$~\citep{Chen2023ParameterEfficientFD} & ICLR, 2023 & LLM & \textcolor{integrated}{H} & \iyes & - & \ino & \iyes & 0.5\% \\ 

NOAH~\cite{zhang2022neural} & TPAMI, 2024 & VFM & \textcolor{integrated}{H} & \iyes & - & \ino & \iyes & 0.4\% \\

DiffFit~\cite{xie2023difffit} & ICCV, 2023 & VGM & \textcolor{integrated}{H} & \ino & - & \iyes & \iyes & 0.12\%\\
\specialrule{.16em}{.4ex}{0pt}
\end{tabular}
\end{adjustbox}
\captionsetup{justification=raggedright, singlelinecheck=false}
\caption{The overview of recent PEFT methods primarily comprises several elements: Approach, Venue, Modal of FMs, Category of PEFT, SC indicates that the structure of FMS has changed, Position denotes the fine-tuned parameter position, IE means inference efficiency, Addition of Parameters, and the percentage of Trainable Parameters. Note that the ``-'' represents that the paper does not provide a clear result. 
}
\label{tab: methods}
\end{table*}

\section{Methodology}
\label{sec: method}
This section will describe several important categories of PEFT methods, encompassing the PEFT taxonomy in LLM, VFM, VLM, MFM, and VGM.
We will also analyze the pros and cons of each category for a more in-depth understanding.

\subsection{\textcolor{selective}{Selective} PEFT}
Methods for this category refer to either selectively fine-tuning a subset of the original model's parameters while keeping the rest frozen, or introducing a minimal number of additional parameters to train, without altering the original parameters, as shown in Table~\ref{tab: comparison_partial}.

\sssection{A.1 Selective PEFT in Basics}

In this group, two core types are included:
specific selection, where predetermined parameters are chosen, and automatic selection, where the model autonomously determines the parameters to be tuned.

\subsubsection{Specific Selection} Methods of this type aim to select specific layers or neurons for fine-tuning. Commonly used methods include \textbf{Freeze Layers}~\cite{Lee2019WhatWE}, \textbf{BitFit}~\cite{BenZaken2021BitFitSP}, and \textbf{PASTA}~\cite{Yang2022ParameterEfficientTW}.

\textbf{Freeze Layers}-based methods only fine-tune the last few layers of FMs inspired by this work~\cite{Kovaleva2019RevealingTD}. \textbf{BitFit}~\cite{BenZaken2021BitFitSP} proposed an even simpler fine-tuning approach by only adjusting part of the bias terms of a model or adjusting all the bias terms of the model. We present the core formula~\eqref{eq: BitFit} for BitFit from the cited paper.
Taking the BERT model as an example, the BERT encoder consists of \textit{L} layers. Each layer begins with \textit{M} self-attention heads, where a self-attention head (\textit{m}, $\ell$) comprises query ($Q$), key ($K$), and value ($V$) encoders, each implemented as a linear layer. Here, $x$ represents the output of the previous encoder layer (for the first encoder layer, $x$ corresponds to the output of the embedding layer). The subject of blue vectors is the {bias terms}.
\begin{align}
\label{eq: BitFit}
Q^{m,\ell}(x) = {W_{q}^{m,\ell}} x + \textcolor{blue}{b_{q}^{m,\ell}}, \nonumber \\
K^{m,\ell}(x) = {W_{k}^{m,\ell}} x + \textcolor{blue}{b_{k}^{m,\ell}}, \\
V^{m,\ell}(x) = {W_{v}^{m,\ell}} x + \textcolor{blue}{b_{v}^{m,\ell}}. \nonumber
\end{align}

\textbf{PASTA}~\cite{Yang2022ParameterEfficientTW} updates only special tokens (e.g., [SEP] and [CLS]), achieving performance similar to full fine-tuning in natural language understanding tasks while training just 0. 029\% of the total parameters. 
In particular, {PASTA}~\cite{Yang2022ParameterEfficientTW} with ${\mbox{RoBERTa}}$ performed similarly to {BitFit}~\cite{BenZaken2021BitFitSP} but with significantly fewer trainable parameters, demonstrating its efficiency. Moreover, on the CoNLL2003~\cite{Sang2003IntroductionTT} for the Named Entity Recognition task, {PASTA}~\cite{Yang2022ParameterEfficientTW} with ${\mbox{RoBERTa}}$ achieved an impressive F1 score of 90.8\%, outperforming {P-tuning} v2~\cite{Liu2021PTuningVP} by 0.6\% with 20 times fewer trainable parameters, although it trailed slightly behind full fine-tuning by 2.0\%. 

\subsubsection{Automatic Selection} Methods of this type aim to utilize various algorithms to determine which parameters to train automatically, such as \textbf{Masking}~\cite{Zhao2020MaskingAA}, \textbf{Diff-Pruning}~\cite{Xu2021RaiseAC}, \textbf{FISH}~\cite{Sung2021TrainingNN, Ansell2021ComposableSF}, \textbf{AutoFreeze Layers}~\cite{Liu2021AutoFreezeAF}, and \textbf{CHILD-TUNING}~\cite{Xu2021RaiseAC}. Compared to specific selection, enabling FMs to decide which parameters to train would be a more sensible and flexible approach.

Inspired by weight agnostic neural networks~\cite{Gaier2019WeightAN}, \textbf{Masking}-based method~\cite{Zhao2020MaskingAA} utilizes the straight-through estimator to train binary masks which are then employed to mask FM's parameters selectively. Zhao et al.~\cite{Zhao2020MaskingAA} show that end-to-end learning of these selective masks, applied both to FMs and a randomly initialized classifier layer, consistently yields excellent performance. It offers a significant advantage in terms of memory footprint, especially when dealing with multiple tasks that need to be inferred simultaneously. Similarly, \textbf{Diff-Pruning}~\cite{Xu2021RaiseAC} learns a specific binary task to explore how FMs can be effectively employed for multitasking in environments with limited storage resources. This method leverages the Diff-vector approach to fine-tune the initial pre-trained parameters which will be separated into two parts, the fixed and the tunable. 
This Diff-vector undergoes adaptive pruning through L0-norm ~\cite{Louizos2017LearningSN} regularization to encourage sparsity. 
However, utilizing Diff-Pruning requires more memories to store the binary mask. \textbf{FISH}-based methods~\cite{Sung2021TrainingNN, Ansell2021ComposableSF} contend that the parameters impacting the final model output constitute only a subset of all parameters. To identify this subset, they create the FISH (Fisher-Induced Sparse uncHanging) Mask, and select top-k parameters with the highest Fisher information as the crucial update parameter.  
FISH Mask remains fixed parameters over multiple iterations, effectively designating a specific subset for modification. Only activated neurons undergo updates during training, while other neurons are masked out. 

\textbf{AutoFreeze Layers}~\cite{Liu2021AutoFreezeAF} utilizes two main modules to accelerate fine-tuning of the model and maintain accuracy: the freezing module and the caching module. 
The freezing module incorporates a decision engine plug-in to determine which layers should be frozen during the training process, employing a gradient-norm test algorithm. 
The caching module aims to store intermediate output results for each layer.
\begin{table}[t!]
    \centering
    \renewcommand{\arraystretch}{1.0}
    \begin{adjustbox}{max width=\linewidth}
    \begin{tabular}{c|c|c|c|c}
        \specialrule{.16em}{0pt} {.65ex}
        \textbf{Selective} & \textbf{Method} & \textbf{Fine-tuning Selection} & \textbf{Backprop} & \textbf{Inference Overhead} \\
        \midrule
        \multirowcell{3}{Specific \\ Selection} & Freeze Layers & \makebox[20ex][r]{The last few layers} & \iyes & \ino \\
         & BitFit & \makebox[20ex][r]{All/selective bias terms} & \ino & \ino  \\
        & PASTA & \makebox[20ex][r]{Special tokens} & \ino &\ino \\
        \cmidrule(lr){1-5}
        \multirowcell{5}{Automatic \\ Selection} & Masking & \makebox[20ex][r]{Binary masks} & \ino & \ino \\
        & AutoFreeze Layer & \makebox[20ex][r]{Freeze layers\&cache} & \ino&\ino \\
        & Diff-Pruning & \makebox[20ex][r]{Nonzero positions} & \ino & \ino \\
        & CHILD-TUNING & \makebox[20ex][r]{Child network} &\ino & \ino\\
        & FISH & \makebox[20ex][r]{Fisher information} & \ino & \ino \\
        \specialrule{.16em}{.4ex}{0pt}
    \end{tabular}
    \end{adjustbox}
    \caption{Key comparison between selective PEFT. Backprop denotes whether reducing backpropagation costs. For Inference Overhead, $\ino$ means there is no extra overhead.}
    \label{tab: comparison_partial}
\end{table}
In contrast, \textbf{CHILD-TUNING}~\cite{Xu2021RaiseAC} identifies a child network in the parameter matrix based on a certain strategy and generates a corresponding mask matrix. 
After computing the gradients, only the parameters corresponding to the child network are updated based on the mask, while the other parameters remain unchanged.
The formula from the cited paper~\cite{Xu2021RaiseAC} is:
\begin{equation}
w_{t+1} = w_{t}-\eta\frac{\partial \mathcal{L}(w_{t})}{\partial w_{t}} \odot M_{t},
\end{equation}
where \textit{t} represents the \textit{t}-th iteration, $w$ is the parameter, $\mathcal{L}$ denotes the loss, ${\eta}$ is the learning rate, and $M_{t}$ signifies the mask matrix. Within $M_{t}$, $1$ indicates that the parameter belongs to the child network, while a value of $0$ means it does not belong to this network.

\sssection{A.2 Selective PEFT in More FMs}

Linear Probe~\cite{radford2021learning} presents CLIP that jointly trains a text encoder and an image encoder enabling zero-shot prediction at test time. 
FC-CLIP~\cite{yu2024convolutions} uses a shared frozen convolutional CLIP backbone to build a single-stage system for open-vocabulary segmentation and consists of three main components (class-agnostic mask generator, in-vocabulary
classifier, and out-of-vocabulary classifier).
Specifically, the classification score can be described as follows:
\begin{equation}
\hat{c}_{i}(j) = \begin{cases}
    (\hat{c}_{i, in}(j))^{(1-\alpha)} \cdot (\hat{c}_{i, out}(j))^\alpha, & \text{if } j \in C_{train}\\
    (\hat{c}_{i, in}(j))^{(1-\beta)} \cdot (\hat{c}_{i, out}(j))^\beta, & \text{otherwise }
\end{cases}
\end{equation}
here, $\hat{c}_i(j)$ represents the $j$-th element of $\hat{c}_i$, with ``in" and ``out" denoting classifiers for in-vocabulary and out-of-vocabulary, respectively. The parameters $\alpha$ and $\beta$, falling within the range [0, 1], control the predictions' balance between the in-vocabulary and out-of-vocabulary classifiers for known and newly unseen categories.
Tune-A-Video~\cite{wu2023tune} presents a text-video pair tuning and proposes a tailored spatiotemporal attention mechanism for text-to-video generations.
\begin{equation*}
\mathcal{V^\ast} = \mathcal{D}(\text{DDIM-samp}(\text{DDIM-inv}(\mathcal{E}(\mathcal{V})), \mathcal{T^\ast})),
\end{equation*}
where $\mathcal{V}$ is a source video, $\mathcal{T^\ast}$ is an edited prompt, and $\mathcal{V^\ast}$ is an output video. During inference, Tune-A-Video leverages DDIM (Denoising Diffusion Implicit Model) inversion to provide structure guidance for sampling from input video $\mathcal{V}$.
LayerNorm Tuning~\cite{zhao2023tuning} only adjusts the weights of the normalization layers within an attention block and demonstrates significant reductions in GPU memory usage and trainable parameters.

\sssection{A.3 Pros and Cons}

Here, we analyze the pros and cons of selective PEFT.
A standout advantage of this category is that it refrains from adding new parameters, which has a dual benefit.
First, it controls the model's complexity by keeping the parameter count in check, preserving the model's manageability. Second, it ensures that the inference time for downstream tasks does not inflate, thus aiding in maintaining model efficiency. Nevertheless, some shortcomings also should be noted.

{\bf \(\bullet\) Memory risk.} Some techniques within this category (like FISH, and CHILD-TUNING), involve the integration of a masking matrix, which results in a spike in memory usage, which could be a challenge in memory-constrained scenarios.

{\bf \(\bullet\) Extra time costs.} Some methods might lead to longer training periods due to a special selection mechanism (like Diff-Pruning). This could potentially offset the benefits of having fewer trainable parameters.

\subsection{\textcolor{additive}{Additive} PEFT}
\begin{figure*}[t!]
  \centering
  \includegraphics[width=1.\textwidth]{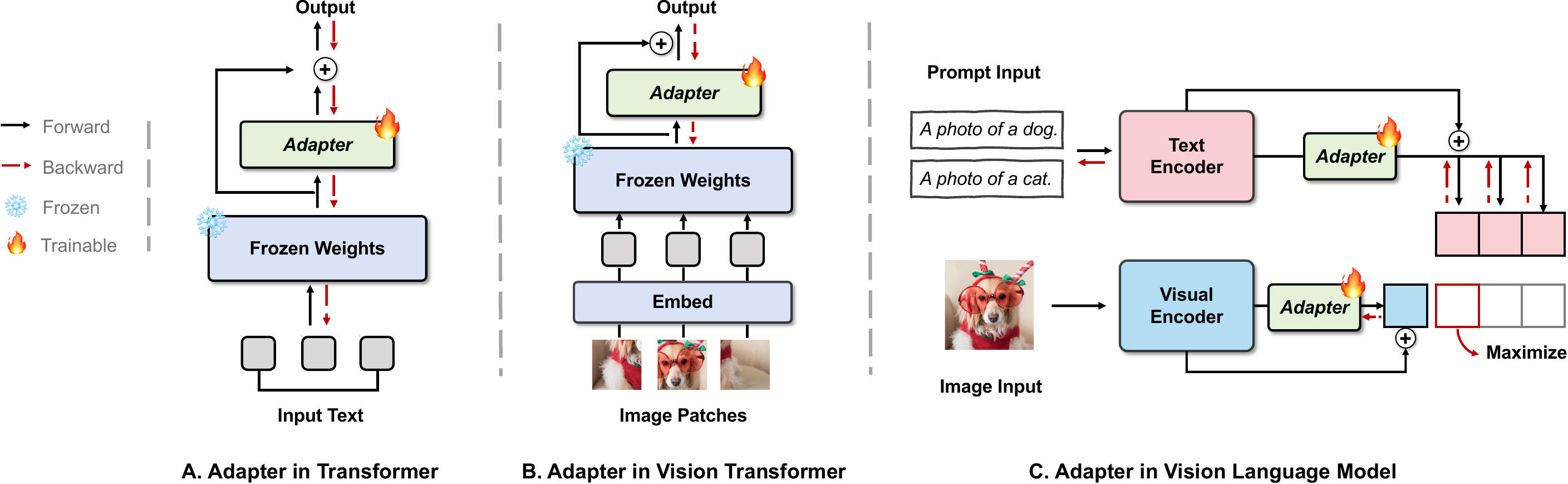}
    \captionsetup{justification=raggedright, singlelinecheck=false}
  \caption{Illustration of representative adapter PEFT across various FMs.}
  \label{fig: adapter}
\end{figure*}

As shown in Fig.~\ref{fig: adapter}, the core idea behind adapters is to learn a set of parameters that can transform the output of one layer into the input of the next layer in a given task-specific way.
Adapters are small parameter sets that can be inserted between the layers of FMs. They allow the network to be fine-tuned for a new task without modifying its original parameters.

\sssection{B.1 Additive PEFT in Basics}

For this group, three key types are included: Bottleneck Adapter, Multi-Adapter, and Adapter Sparsity, as shown in Table~\ref{tab: comparison_additive}. 

\subsubsection{Bottleneck Adapter} Method of this type~\cite{Houlsby2019ParameterEfficientTL} is proposed in the NLP field inspired by Residual Adapter~\cite{Rebuffi2017LearningMV} and ResNet~\cite{He2016IdentityMI} in cross-domain image classification task. This work~\cite{Houlsby2019ParameterEfficientTL} demonstrates the feasibility of using adapters for parameter-efficient transfer learning on classic NLP tasks.
The adapter layer has a simple structure: it is down-projected to a smaller dimension, passed through a non-linear activation function, and then up-projected to the original dimension like a bottleneck. In addition, there is a residual connection between the input and output of the entire adapter layer.
However, due to the additional parameters introduced by adapters, the model's inference speed has slowed, leading to various pruning operations on adapters. Further, how to make adapters lighter without sacrificing their performance has become a hot research direction.

\subsubsection{Multi-Adapter} Methods of this type refer to the addition of more adapter modules to the model to enhance its transferability. These methods are proposed as a specialized knowledge plug-in to integrate the knowledge of various tasks without forgetting the knowledge from previous tasks and improve the performance of the Bottleneck Adapter~\cite{Houlsby2019ParameterEfficientTL}.
Multi-Adapter mainly includes Adapter Fusion~\cite{Pfeiffer2020AdapterFusionNT}, AdaMix~\cite{Wang2022AdaMixMF}, MAD-X~\cite{Pfeiffer2020MADXAA}, and BAD-X~\cite{Parovic2022BADXBA}.

\textbf{Adapter Fusion}~\cite{Pfeiffer2020AdapterFusionNT} combines the knowledge from multiple tasks by fusing the parameters of their respective adapters.
This multi-task learning framework consists of two stages. First, a set of new adapter parameters are learned for each task. Then, a fusion module is learned for a specific target task to combine all adapters learned from the first stage. It is worth noting that this method does not require any changes to the structure or parameters of the adapters, but rather combines multiple adapters through simple addition, making it a non-destructive approach.
\textbf{AdaMix}~\cite{Wang2022AdaMixMF} does a similar thing as Adapter Fusion~\cite{Pfeiffer2020AdapterFusionNT}, by reconstructing the structure of the adapter. 
Given the output of the expert $\mathbb{E}_{i}$:
\begin{equation}
\mathbb{E}_{i}\left(x_{s}\right)=w_{i}^{\text {out }} \cdot \text{GeLU}\left(w_{i}^{\text {in }} \cdot x_{s}\right),
\end{equation}
where $x_s$ is the input token representation at the $s$-th position for the MoE (Mixture-of-Expert) layer, consisting of \(N\) expert Feed-Forward Network (FFN, $\mathbb{E}_{i}$), \(w_{i}^{\text{{in}}}\) and \(w_{i}^{\text{{out}}}\) denote the input and output projection matrices for the \(i\)-th expert. The Softmax function is GeLU~\cite{Hendrycks2016GaussianEL}.
Using a gating network, the output of MoE is given by:
\begin{equation}
h(x_s) = \sum_{i} \mathbb{G}(x_s)_i \mathbb{E}_i(x_s),
\end{equation}
where $\mathbb{G}(x_s)_i$ is the $i\text{-th}$ logit of the output of $\mathbb{G}(x_s)$, which denotes the probability of selecting expert $\mathbb{E}_i$.
Subsequently, Wang et al.~\cite{Wang2022AdaMixMF} replace the gating unit with a random average selection methodology about experts serves not only to mitigate the computational load and the number of parameters necessitated by the gating unit but also to avert the risk of overload for any individual expert. However, AdaMix demands more memory resources during the training process.
 
\textbf{MAD-X} (Multiple ADapters for Cross-lingual transfer) framework~\cite{Pfeiffer2020MADXAA} comprises three types of adapters: invertible, language, and task adapters. {Invertible adapters} are modules added on top of the embedding layer. The inverse comes before the output embedding layer. This setup helps tackle vocabulary mismatches between multilingual and target languages. {Language adapters} are trained using masked language modeling on unlabeled data for a specific language. This training encourages the adapters to learn transformations that enhance the pre-trained multilingual model's suitability for that particular language. {Task adapters} are designed to learn specific tasks. When updating the parameters of the task adapter, the language adapter and inverse adapter are frozen. MAD-X is particularly useful for adapting to languages not covered by the multilingual model’s training model and achieving competitive performance on high-resource languages.
\textbf{BAD-X}~\cite{Parovic2022BADXBA} advocates for a more effective cross-lingual transfer by directly adapting the model to the specific source-target language pair, rather than separately training the source language and target language through a monolingual adapter. In this approach, a bilingual language pair adapter is learned, optimizing the adaptation process and potentially improving cross-lingual performance.
Even though Multi-Adapter enhances transferability, this implementation introduces more parameters.

\subsubsection{Adapter Sparsity} Methods of this type are proposed to make full use of the parameter efficiency according to the internal structure of the adapter. Like AdapterDrop~\cite{Rckl2020AdapterDropOT}, LST~\cite{Sung2022LSTLS}, and Convpass~\cite{jie2022convolutional}.

\textbf{AdapterDrop}~\cite{Rckl2020AdapterDropOT} aims to reduce computation and memory requirements and simplify the model. They achieve this by randomly dropping adapters during training, which encourages the model to learn to rely on the original transformer layers in addition to the adapters. This can result in faster and more efficient training. 
AdapterDrop proposed two training approaches: (1) specialized adapter dropout, wherein during training, a fixed value $n$ is maintained, and the model retains only the top $n$ layers during inference. (2) robust AdapterDrop that randomly draws $n$ from [0, 11] for training. Across multiple tasks in the GLUE benchmark, both variants of AdapterDrop demonstrate minimal performance degradation when dropping below $n$=5 during inference, whereas traditional adapters experience a rapid decline in performance beyond $n$=1. Removing the top 5 layers of the adapter results in a 26\% increase in training speed. Moreover, the speed of concurrent inference for multiple tasks can be accelerated by 21\%-42\%.
\textbf{AdapterBias}~\cite{Fu2022AdapterBiasPT} incorporates the idea of BitFit~\cite{BenZaken2021BitFitSP} and introduces a token-dependent shift to the hidden output of transformer layers to adapt to downstream NLP tasks with only a vector and a linear layer, and demonstrates its efficiency by employing BERT as the pre-trained language model. Compared to Bottleneck Adapter~\cite{Houlsby2019ParameterEfficientTL}, AdapterBias shows competitiveness despite having 40 times fewer parameters.

\textbf{SparseAdapter}~\cite{He2022SparseAdapterAE} further checks the additive PEFT from the perspective of network pruning and introduces the concept of Large-Sparse to maintain the same parameter budget. SparseAdapter can achieve comparable, or even better, performance than standard adapters when the sparsity ratio reaches 80\%.
\textbf{LST}~\cite{Sung2022LSTLS}, as a variant of adapters, involves training a small transformer network on one side of a pre-trained network, similar to a ladder. This network is connected to the transformer layers of the original model. FMs are utilized solely as feature extractors, while backpropagation is performed within the side network. LST~\cite{Sung2022LSTLS} employs various techniques to conserve memory and computational resources during training and enhance fine-tuning performance.

\begin{table}[t!]
    \centering
    \renewcommand{\arraystretch}{1.0}
    \begin{adjustbox}{max width=\linewidth}
    \begin{tabular}{c|c|c|c|c}
        \specialrule{.16em}{0pt} {.65ex}
        \textbf{Additive} & \textbf{Method} & \textbf{Fine-tuning Selection} & \textbf{Backprop} & \textbf{Inference Overhead} \\
        \midrule
        \multicolumn{2}{c|}{\multirowcell{2}{Bottleneck Adapter}} & \makebox[35ex][r]{\multirowcell{2}{Down-project\&non-linear activation \\\&up-project}} & \multirowcell{2}{\ino} & \multirowcell{2}{FFN} \\
        \multicolumn{2}{c|}{} & & & \\
        \cmidrule(lr){1-5}
        \multirowcell{4}{Multi-Adapter} & Adapter Fusion & \makebox[35ex][r]{Combine all adapters} & \ino & decoder \\
         & MAD-X  & \makebox[35ex][r]{Invertible, language, and task adapters} &\ino & FFN \\
         & BAD-X & \makebox[35ex][r]{A bilingual language pair adapter} &\ino & FFN \\
         & AdaMix & \makebox[35ex][r]{A MoE layer} & \ino & FFN \\
        \cmidrule(lr){1-5}
        \multirowcell{4}{Adapter Sparsity} & AdapterDrop & \makebox[35ex][r]{Randomly dropping adapters} &\ino & FFN\\
         & AdapterBias & \makebox[35ex][r]{A token-dependent shift} &\ino & \ino\\
         & SparseAdapter & \makebox[35ex][r]{Pruning network} &\ino & FFN\\
         & LST & \makebox[35ex][r]{Ladder-side tuning} & \iyes & decoder \\
        \specialrule{.16em}{.4ex}{0pt}
    \end{tabular}
    \end{adjustbox}
    \caption{Key comparison between additive PEFT. Backprop denotes whether reducing backpropagation costs. For Inference Overhead, $\ino$ means there is no extra overhead, FFN means adding overhead to the FFN part, and others are similar.}
    \label{tab: comparison_additive}
\end{table}

\sssection{B.2 Additive PEFT in More FMs}

LST~\cite{Sung2022LSTLS} has been evaluated on T5~\cite{Raffel2019ExploringTL} and CLIP-T5~\cite{Sung2021VLADAPTERPT} models, revealing that when fine-tuning the entire network, LST reduces memory costs by 69\%, whereas other methods achieve only a 26\% reduction under similar parameter usage.
Convpass~\cite{jie2022convolutional} introduces convolutional bypasses in ViTs as vision transformer adapters by introducing less than 0.5\% trainable parameters for adapting vision models.
AdaptFormer~\cite{chen2022adaptformer} introduces a lightweight module with less than 2\% of the parameters of the ViT to boost recognition performance.
ViT-Adapter~\cite{chen2022vision} enhances the intrinsic representational capabilities of a standard ViT backbone with an adapter that integrates image-specific inductive biases during the fine-tuning process.
SAN~\cite{xu2023side} separates mask proposal generation and class recognition tasks to achieve open-vocabulary semantic segmentation. By appending a lightweight side network to a fixed CLIP model, mask proposals and attention bias are predicted to direct CLIP in recognizing the class of the mask.
CSN (DTL)~\cite{fu2024dtl} disentangles the weight updates from the backbone using a compact side network to identify the object.
T2I-Adapter~\cite{mou2024t2i} learns lightweight adapter modes $\mathcal{A}$ to improve the performance of text-to-image models without the inherent framework of updating text-to-image models $\mathcal{M}$.
Given text prompt $t$, control signal $x_c$, and weighting factor $w$, T2I-Adapter can generate image $x$:
\begin{equation}
    x = \mathcal{M}(t) + w \cdot \mathcal{A}(x_c).
\end{equation}
IP-Adapter~\cite{ye2023ip} uses an image prompt and introduces a cross-attention mechanism to learn image embeddings effectively. 
Given conditional model $\boldsymbol{\epsilon}_{\theta}(\boldsymbol{x}_t, \boldsymbol{c}, t)$ and unconditional model $\boldsymbol{\epsilon}_{\theta}(\boldsymbol{x}_t, t)$, the predicted noise is:
\begin{equation}
\hat{\boldsymbol{\epsilon}}_{\theta}(\boldsymbol{x}_t, \boldsymbol{c}, t) = w\boldsymbol{\epsilon}_{\theta}(\boldsymbol{x}_t, \boldsymbol{c}, t)+(1-w)\boldsymbol{\epsilon}_{\theta}(\boldsymbol{x}_t, t),
\end{equation}
$w$ is the guidance scale or weight that adjusts the alignment with condition $\boldsymbol{c}$. 
I2V-adapter~\cite{guo2024i2v} needs to fine-tune only 1\% of the parameters of the base diffusion model. 
ControlNet~\cite{zhang2023adding} adds spatially localized conditions. 
Subsequently, ControlNeXt~\cite{peng2024controlnext} introduces a lightweight conditional control module that further reduces learnable parameters to less than 10\% of ControlNet, extending the scope to video generation and super-resolution. 
LLaMA-Adapter V2~\cite{gao2023llama} efficiently enhances LLaMA-Adapter~\cite{zhang2023llama} by unlocking more learnable parameters. 
And CLIP-Adapter~\cite{gao2024clip} and Tip-Adapter~\cite{zhang2021tip}, etc. suggest inserting trainable adapters to perform VLM fine-tuning into the fixed CLIP model. 

\sssection{B.3 Pros and Cons}

Here, we analyze the pros and cons of additive PEFT.
This category integrates task-specific parameters into the model by adding lightweight adapter layers to each layer and does not change most of the weights of FMs, thus preserving the integrity of the pre-trained knowledge.
This makes the adapter model more generic and allows it to leverage rich knowledge to adapt to different tasks without having to retrain the entire model from scratch for each new task. This is especially valuable for rapid deployment and transfer learning scenarios.
Nevertheless, some shortcomings also should be noted.

{\bf \(\bullet\) Inference overhead.} This category may cause an increase in inference overhead due to the additional computation required by the adapter layer.

{\bf \(\bullet\) Prudent configurations.} This category of methods may require careful initialization and training strategies, such as optimal settings of adapter dimensions and sparsity rates.

\begin{figure*}[t!] 
\centering 
\includegraphics[width=1.\textwidth]{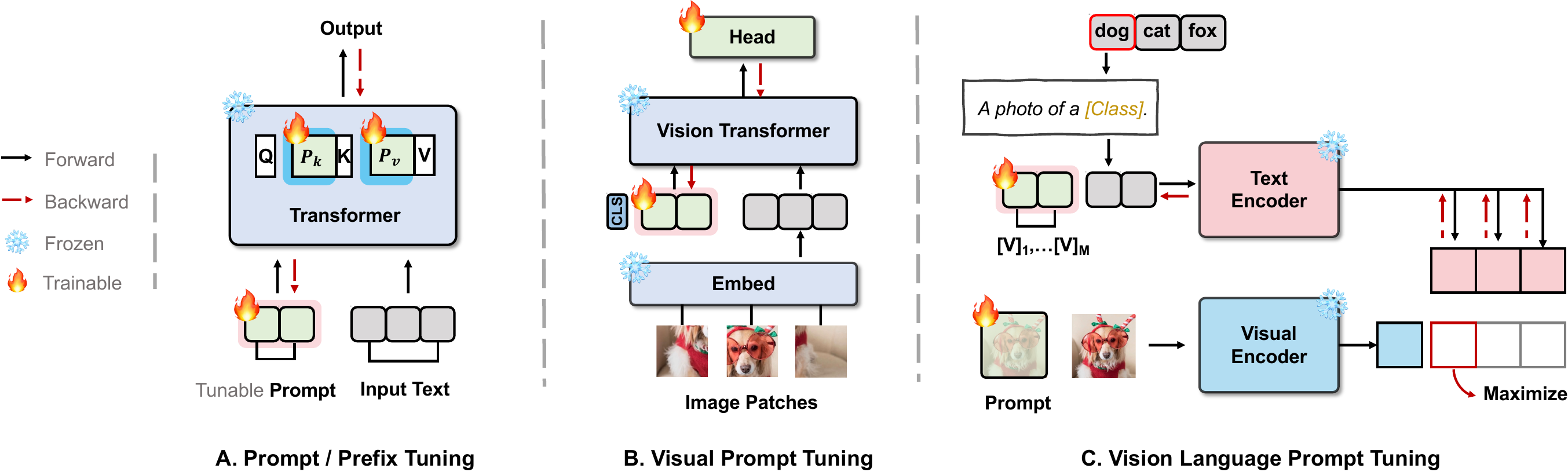}
\captionsetup{justification=raggedright, singlelinecheck=false}
\caption{Illustration of representative prompt PEFT across various FMs.} 
\label{fig: Prompt} 
\end{figure*}

\subsection{\textcolor{prompt}{Prompt} PEFT}

Prompt Tuning is nearly the most common PEFT method in FMs on some specific tasks, as shown in Fig.~\ref{fig: Prompt}.
This category involves incorporating a carefully designed prompt into the input or the transformer's layers, aiming to align the input distribution with the original training data and guide the model toward generating the desired output.

\sssection{C.1 Prompt PEFT in Basics}

Three types are discussed here: Hard Prompt, AutoPrompt, and Soft Prompt.

\subsubsection{Hard Prompt} This type of approach means the initial form of the prompt involves manually specifying a template and concatenating it with the input to generate the desired output, without modifying the original model parameters.

\textbf{PET}~\cite{Schick2020ExploitingCF} proposed pattern-exploiting training, a semi-supervised technique that reformulates input examples into cloze-like sentences. For example, in a task to determine whether two sentences $a$ and $b$ agree or disagree, a pattern like $P(a, b) = a? \underline{\quad}, b$ is used. PET predicts the correct label by filling in the blank.
\textbf{Null Prompts}~\cite{LoganIV2021CuttingDO} uses a general template like ``input + [MASK]'' to simplify the process of designing prompts for downstream tasks and reduce the time spent on prompt engineering and enhances memory efficiency.

Although hard prompts are sometimes efficient, they have notable limitations: (1) selecting effective templates often requires significant human effort, making the process time-consuming and labor-intensive. (2) the generalization ability of hard prompts may suffer when the model encounters new or unfamiliar tasks, which may require further adjustment and pattern learning.

\subsubsection{AutoPrompt} This category proposes an automated prompt search method~\cite{Shin2020ElicitingKF} that uses exploratory search to automatically generate prompts to address the challenges of manual prompt design of Hard Prompt. Although these automatically generated templates may not follow natural language conventions, the terms within them are still understandable, as they are selected from the model's vocabulary. However, the generated templates may not always represent the optimal solution.

\subsubsection{Soft Prompt} This category has further expanded the scope beyond human-understandable words found in a vocabulary. These prompts are called continuous or soft prompts. In this advanced progression, the generation process changes from discrete, human-driven to continuous, machine-driven. Representative methods include Prefix Tuning~\cite{Li2021PrefixTuningOC}, Prompt Tuning~\cite{Lester2021ThePO}, P-Tuning~\cite{Liu2021GPTUT, Liu2021PTuningVP}, PPT~\cite{Gu2021PPTPP}, and so on.

\textbf{Prefix Tuning}~\cite{Li2021PrefixTuningOC} freezes the parameters of FMs and optimizes only a task-specific continuous vector known as the prefix, which functions as a differentiable virtual Token. An MLP is introduced before the prefix layer to enhance stability during training and prevent performance degradation and only the prefix parameters are retained. 
Despite tuning only around 0.1\% of the model's parameters, prefix tuning achieved comparable performance to full fine-tuning on E2E~\cite{Novikova2017TheED}, WebNLG~\cite{Gardent2017TheWC}, and DART~\cite{Radev2020DARTOS} on GPT-2 and BART~\cite{Yuan2022BioBARTPA}.
\textbf{Prompt Tuning}~\cite{Lester2021ThePO} is a simplified version of prefix tuning and defines task-specific prompts, which are appended to the input data. Unlike prefix tuning, prompt tuning only adds prompt tokens to the input layer. Lester et al.~\cite{Lester2021ThePO} also introduced Prompt Ensembling, where multiple prompts for the same task are trained simultaneously within one batch, mimicking model ensembling but at a lower cost. They further investigated how prompt initialization techniques and prompt length impact performance. Their ablation studies show that initializing prompts with class labels is better than other methods like random or vocabulary-based initialization. However, this advantage diminishes as the model size increases. Regarding prompt length, the optimal performance is achieved with about 20 tokens, beyond which no significant improvements are observed. However, this trend also weakens with larger models.

Compared to Prefix Tuning~\cite{Li2021PrefixTuningOC}, \textbf{P-Tuning}~\cite{Liu2021GPTUT} applies differentiable virtual tokens only  at the input layer, rather than across all layers, and allows flexible token insertion rather than restricting them to a prefix position.
Specifically, P-Tuning transforms prompts into a learnable embedding layer, processing them through an MLP and LSTM structure~\cite{Shi2015ConvolutionalLN}. This approach replaces traditional hand-crafted tokens with learnable virtual tokens. Prompt Tuning~\cite{Lester2021ThePO} and P-Tuning only apply prompts to the first transformer layer, resulting in shallow tuning and restricted optimization, especially when applied to smaller models and hard sequence tagging tasks. 
\textbf{P-Tuning v2}~\cite{Liu2021PTuningVP} then extends prompt tokens to each layer of the model, enhancing scalability and universality across various natural language understanding tasks. By incorporating prompts at every layer, P-Tuning v2 increases the number of learnable parameters, from 0.01\% in P-Tuning and Prompt Tuning to 0.1\%-3\%, while maintaining parameter efficiency. This deeper integration improves model predictions and can be seen as an adaptation and enhancement of prefix tuning.

\textbf{DART}~\cite{Zhang2021DifferentiablePM} treats the generation of the prompt as a differentiable function, enabling the model to learn the best way to generate a prompt for a particular task and allowing for gradient-based optimization of the prompt generation. It is similar to P-Tuning~\cite{Liu2021GPTUT} but with some differences in the details, such as using continuous labels, and incorporating a template mask objective instead of an LSTM in P-Tuning.
$\mathcal{\textbf{y}}$\textbf{-Tuning}~\cite{Liu2022YTuningAE} fine-tunes the label extractor's parameters and uses cross attention to combine the loss features from both FMs and the label extractor to avoid adjusting the input text attributes or the parameters of FMs.

\begin{table}[t!]
    \centering
    \renewcommand{\arraystretch}{1.0}
    \begin{adjustbox}{max width=\linewidth}
    \begin{tabular}{c|c|c|c|c}
        \specialrule{.16em}{0pt} {.65ex}
        \textbf{Prompt} & \textbf{Method} & \textbf{Strategy} & \textbf{Backprop} & \textbf{Inference Overhead} \\
        \midrule
        \multirowcell{2}{Hard \\ Prompt} & PET & \makebox[25ex][r]{Reformulate inputs} & \iyes& input \\
        & Null Prompts & \makebox[25ex][r]{Input + [MASK]} & \iyes   & input  \\
        \cmidrule(lr){1-5}
        \multicolumn{2}{c|}{\multirow{2}{*}{Auto-Prompt}} & \makebox[25ex][r]{\multirowcell{2}{Automatically generate \\ Optimal prompts}} & \multirowcell{2}{\iyes} & \multirowcell{2}{input} \\
        \multicolumn{2}{c|}{} & & & \\ 
        \cmidrule(lr){1-5}
        \multirowcell{8}{Soft\\ Prompt} & Prefix Tuning & \makebox[25ex][r]{Train task-oriented prefix} & \ino & input\\
        & Prompt Tuning & \makebox[25ex][r]{Prefix only at input layer} & \ino & input\\
        & P-tuning & \makebox[25ex][r]{Learnable prompt layer} &\ino &input \\
        & DART & \makebox[25ex][r]{Learnable masked prompt} &\ino &input \\
        & \textit{y}-tuning & \makebox[25ex][r]{Train label extractor} & \ino&FFN \\
        &PPT & \makebox[25ex][r]{Pre-train prompt} & \ino & input\\
        & SPoT & \makebox[25ex][r]{Pre-train multi-task prompt} & \ino & input\\
        & Prompt Transfer & \makebox[25ex][r]{Utilize trained soft prompts} & \ino&input \\
        \specialrule{.16em}{.4ex}{0pt}
    \end{tabular}
    \end{adjustbox}
    \caption{Key comparison between prompt PEFT. Backprop denotes whether reducing backpropagation costs. For Inference Overhead, input means adding overhead to the input part, and others are similar.}
    \label{tab:my_label_prompt_tuning}
\end{table}

\textbf{PPT}~\cite{Gu2021PPTPP} propose Pre-trained Prompt Tuning, where soft prompts are pre-trained on a large-scale, unlabeled corpus through self-supervised tasks.
PPT involves two steps: pre-training and fine-tuning. During pre-training, a large dataset is used for self-supervised learning to generate universal prompts. In the fine-tuning stage, generated prompts and a small amount of labeled data are used to fine-tune models on target tasks. PPT excels in few-shot settings, outperforming prompt tuning~\cite{Lester2021ThePO}, with significant improvements in LCQMC~\cite{Liu2018LCQMCALC}, and comparable results in BoolQ~\cite{Clark2019BoolQET}. However, its advantage diminishes with more training data.
\textbf{SPoT}~\cite{Vu2021SPoTBF} shares similarities with PPT, using pre-trained prompts to enhance few-shot learning. Instead of manually designing pre-training tasks, SPoT initializes target task prompts using those trained on source tasks. SPoT inserts a pre-training step between LLM pre-training and target task prompt tuning, training one or multiple prompts on source tasks before using them to initialize prompts for the target task.
SPoT focuses on full-data scenarios, finding that even with sufficient data, using source task-trained prompts provides significant benefits for the target task. 
\textbf{Prompt Transfer}~\cite{Su2021OnTO} involves reusing trained soft prompts for zero-shot inference on new tasks or datasets, or for continued training, showing that soft prompts are effective for similar tasks under the same FMs. Transferability across models is achieved using a cross-model projector. In addition, using these prompts as a starting point for new tasks reduces training time and enhances performance. Performance metrics, particularly the overlapping rate of activated neurons, suggest that prompts stimulate the inherent abilities of FMs.

\sssection{C.2 Prompt PEFT in More FMs}

VP~\cite{bahng2022exploring} adapts FMs to new tasks by adding prompts in the form of pixels to the image’s pixel space, such as padding pixels along the image edges, without altering the model’s parameters. VPT~\cite{jia2022visual} then introduces some learnable parameters in the input space that are less than 1\% of the original model parameters.
DAM-VP~\cite{huang2023diversity} enhances the performance of pre-trained models on downstream tasks with high diversity and large datasets by adaptively selecting and optimizing visual prompts for different subsets of images.
Given input image $x^p$ and ground truth $y$, the the cross-entropy loss on a dataset $\mathcal{D}_\mathcal{T}$ that includes prompts $p_k$ is: 
\begin{equation}
    p_1^\ast, \dots, p_N^\ast = \mathop{\arg\min}_{p_1, \dots, p_N} \frac{1}{|\mathcal{D}_\mathcal{T}|} \sum_{i=1}^N \sum_{x\in \mathcal{D}_i} \mathcal{L}_{CE} (\mathcal{M}(x + p_i), y). 
\end{equation}
ILM-VP~\cite{chen2023understanding} advances visual prompting in transfer learning by introducing an iterative label mapping-based framework that significantly improves the precision of the target task and outperforms existing methods.
EVP~\cite{wu2022unleashing} significantly improves classification accuracy on various datasets to 82.5\% by treating prompts as learnable entities and applying input diversity with gradient normalization, surpassing previous records.
LION~\cite{wang2024lion} is a lightweight and effective vision prompt tuning method that leverages implicit equilibrium layers to adapt pre-trained models to downstream tasks with minimal computational cost.
Textual Inversion~\cite{gal2022image} found a way to describe novel concepts in the text encoder of CLIP to fine-tune the diffusion model (using less than 20k parameters) to generate content in a specialized style.
CoOp~\cite{zhou2022learning} models the context words of the prompt with learnable vectors for implementing PEFT to identify or detect objects.
OVSeg~\cite{liang2023open} incorporates masked and colorful prompts to improve the fine-tuning performance of VFMs significantly. 
Q-Former~\cite{li2023blip} bridges the modal gap using a lightweight projection that greatly reduces trainable parameters.

\sssection{C.3 Pros and Cons}

Here, we analyze the pros and cons of prompt PEFT. This category of PEFT adjusts the corresponding learnable prompt vectors (like text prompt and visual prompt) while maintaining consistent architecture, greatly improving the flexibility and versatility of the model. Second, since the base model parameters remain fixed, it helps preserve knowledge across tasks, reducing forgetting in multi-task scenarios.
Nevertheless, some shortcomings also should be noted.

{\bf \(\bullet\) Poor Transferability.} Some prompts trained for specific a task cannot be directly transferred to other tasks. Because the prompt vectors for each task are optimized based on the data and features of that task, they have strong task-specific characteristics and are not easily generalized across different tasks.

{\bf \(\bullet\) Model Dependency.} This category of PEFT relies on the model's already possessed capabilities. If the FMs have some deficiencies, it is difficult to compensate for these shortcomings through prompt tuning, and the room for performance improvement is limited.

\subsection{\textcolor{reparameter}{Reparameterization} PEFT}
\begin{figure}[t!] 
\centering 
\includegraphics[width=0.49\textwidth]{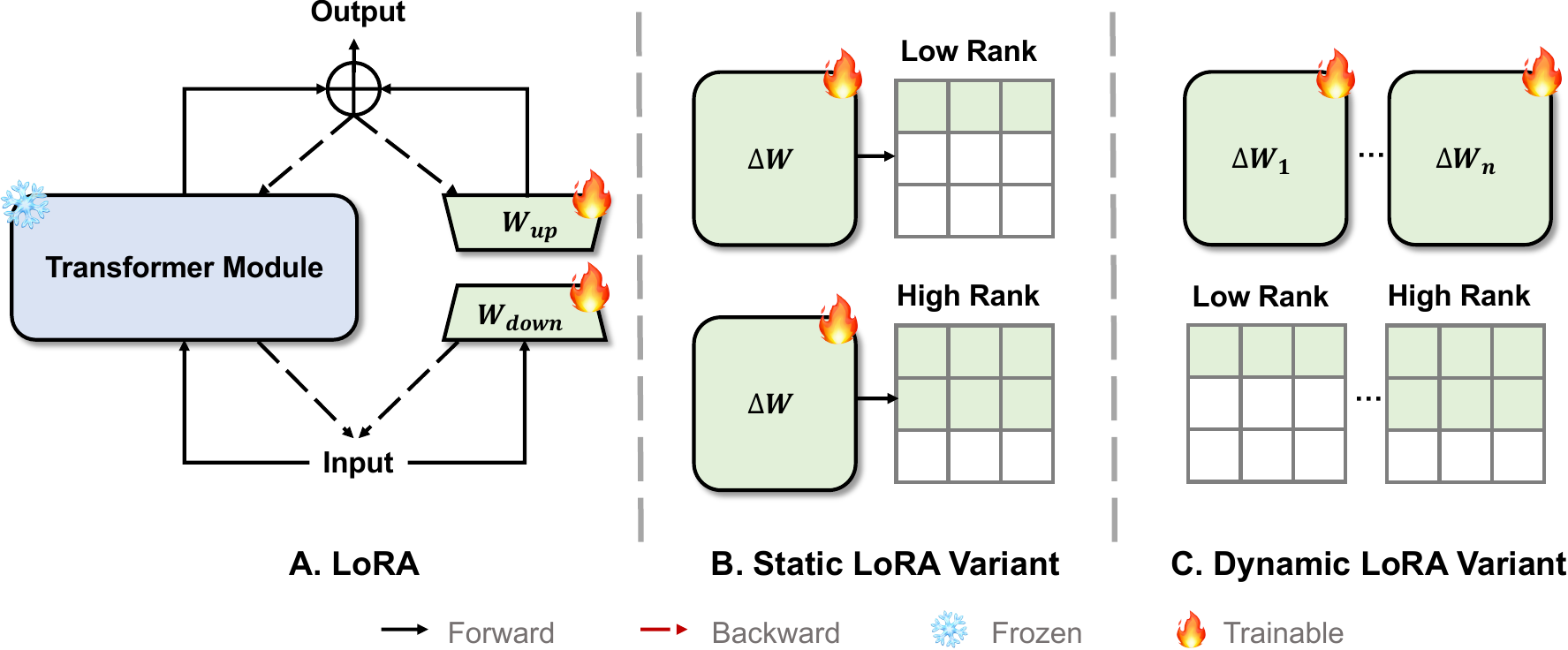}
    \caption{Illustration of representative groups of LoRA PEFT.}
    \label{fig: diff_lora}
\end{figure}

While the additive PEFT reduces the number of tunable parameters by employing down-project and up-project techniques, its synthetic structure can negatively impact the model's inference speed. Similarly, training the prompt in prompt tuning may be unstable, as it relies on human input, which is often subjective. Additionally, including prompt tokens in the input sequence can reduce the effective sequence length, potentially leading to suboptimal performance. To address these limitations, we introduce another PEFT technique, Reparameterization, as shown in Fig.~\ref{fig: diff_lora} and Table~\ref{tab: comparison_reparameterization}.
This technique reparameterizes the low-dimensional representation of the initial model parameters for training while converting the weights back for inference.

\sssection{D.1 Reparameterization PEFT in Basics}

Reparameterization mainly includes two groups: LoRA~\cite{Hu2021LoRALA} and its variants~\cite{Dettmers2023QLoRAEF, zhang2023lora}, and MPO~\cite{Liu2021EnablingLF}.

\subsubsection{LoRA and Its Variants} LoRA capitalizes on the low-rank structure inherent in many machine learning problems~\cite{Li2016RecoveryGO, Li2017AlgorithmicRI, Oymak2019GeneralizationGF} as a basic reparameterization technique. Aghajanyan et al.~\cite{Aghajanyan2020IntrinsicDE} delve into the intrinsic dimensionality and demonstrate that natural language tasks can be tackled with a surprisingly small number of parameters, sometimes only a few hundred. This discovery implies that the pre-training of FMs can be regarded as a form of knowledge compression, where each task corresponds to a unique intrinsic dimension within the model's subspace. Empirical studies indicate that larger models tend to have lower intrinsic dimensions than their baseline counterparts.

\textbf{LoRA}~\cite{Hu2021LoRALA} is a pioneering method that explores low-rank updates for adapting a fixed model to downstream tasks. LoRA's simplicity lies in its addition of a bypass to FMs, which includes a self-attention module and performs rank manipulation to approximate the intrinsic rank. During training, only the low-rank matrix $\mathbb{A}$ and high-rank matrix $\mathbb{B}$ are updated, with $\mathbb{A}$ initialized randomly and $\mathbb{B}$ as a zero matrix, ensuring the bypass starts with no effect.
\begin{equation}
h=W_{o}x+\bigtriangleup Wx= \bigtriangleup W_{o}x+\mathbb{B}\mathbb{A}x.
\end{equation}
The model's input and output dimensions are preserved. LoRA enhances the output by adding the product of matrices $\mathbb{B}$ and $\mathbb{A}$ to FMs' parameters. \textbf{KronA}~\cite{Edalati2022KronAPE} differs from LoRA by using Kronecker products instead of low-rank matrices, offering greater expressivity. KronA excels at capturing complex data relationships.
\textbf{QLoRA}~\cite{Dettmers2023QLoRAEF} was introduced to enable fine-tuning of FMs, such as a 65-billion-parameter model, on smaller GPUs (e.g., a single 48GB GPU) while maintaining full 16-bit fine-tuning performance. QLoRA achieves this by using frozen 4-bit quantized FMs, which propagate gradients to LoRA during backpropagation. QLoRA introduces several memory-saving techniques without compromising performance, such as a new 4-bit NormalFloat data type optimized for weight distributions, double quantization to reduce memory usage through quantization constants, and a paging optimizer to manage memory peak values effectively.

Unlike the basic LoRA method, \textbf{LoRA-FA}~\cite{zhang2023lora} is a memory-efficient fine-tuning method designed to reduce activation memory needs. LoRA-FA freezes the projection-down weight of matrix $\mathbb{A}$ and updates only the projection-up weight of matrix $\mathbb{B}$ in each LoRA layer. This keeps model weight changes within a low-rank space, eliminating the need to store full-rank input activations. Experiments across models show that LoRA-FA achieves accuracy comparable to full fine-tuning while reducing memory costs by up to 1.4 times compared to standard LoRA.
\textbf{IncreLoRA}~\cite{zhang2023increlora} improves LoRA by dynamically adding trainable parameters based on module importance scores. This allows for better parameter efficiency, especially in low-resource scenarios, without being limited by an initial parameter count.
\textbf{Delta-LoRA}~\cite{Zi2023DeltaLoRAFH} not only updates the low-rank matrices $\mathbb{A}$ and $\mathbb{B}$ but also adjusts the pre-trained weights $W$ through the differential of their product $(\mathbb{A}(t+1)\mathbb{B}(t+1)-\mathbb{A}(t)\mathbb{B}(t))$. This approach effectively overcomes the limitations of incremental updates in low-rank matrices, enhancing the model's ability to learn task-specific representations. Note that Delta-LoRA realizes these advancements without a substantial increase in memory usage or computational expense compared to LoRA.

\subsubsection{MPO} The matrix product operator is a representation of tensor networks characterized by slow growth in parameters and computational complexity with increasing input dimensions, making them suitable for compressing FMs. The MPO~\cite{Liu2021EnablingLF} decomposes the parameter matrix and defines the central tensor and auxiliary tensors. Given the nature of MPO decomposition, the central tensor contains significantly more parameters than the auxiliary tensors, suggesting it encapsulates the essential linguistic information of FMs. For downstream task adaptation, only the low-parameter auxiliary tensors need training.

\sssection{D.2 Reparameterization PEFT in More FMs}

LoRand~\cite{yin20231} leverages low-rank decomposition to create compact adapters for fine-tuning, achieving competitive performance with just 1-3\% of the original model's parameters, significantly reducing the computational overhead.
LyCORIS~\cite{yeh2023navigating} provides an advanced suite of tools for fine-tuning Stable Diffusion models, enhancing their capabilities for text-to-image generation with improved control and quality.
DiffuseKronA~\cite{marjit2024diffusekrona} employs Kronecker product decomposition to minimize parameters in attention layers of diffusion models, achieving substantial efficiency gains without compromising image generation quality.
Mix-of-Show~\cite{gu2024mix} proposes embedding-decomposed LoRA (ED-LoRA) to train a single concept, gradient fusion for the center-node concept fusion, and regionally controllable sampling for diffusion models.
LoRA-Sparse~\cite{song2024low} develops low-rank linear projection layers for sparse attention to enhance the performance of LLaVA-1.5.

\sssection{D.3 Pros and Cons}

Here, we analyze the pros and cons of reparameterization PEFT. This category of PEFT features high flexibility. Such as it can be applied to almost all mainstream models and is very flexible, allowing for rapid adaptation to new tasks and domains. Whether it's LLMs like GPT or VGMs like stable diffusion, LoRA can be easily used for fine-tuning to adapt to different application scenarios and task requirements.
Nevertheless, some shortcomings also should be noted.

{\bf \(\bullet\) Hyperparameters sensitivity.} This type of method is sensitive to hyperparameters. Like, the rank of the inserted adaptation matrices significantly impacts the ability to adapt the model to a new task.

{\bf \(\bullet\) Limited representation.} This category of PEFT assumes that model adaptations can be represented using low-rank matrices. In tasks where the feature space is highly complex, this assumption may limit expressiveness and lead to suboptimal performance.

\begin{table}[t!]
    \centering
    \renewcommand{\arraystretch}{1.0}
    \begin{adjustbox}{max width=\linewidth}
    \begin{tabular}{c|c|c|c|c}
        \specialrule{.16em}{0pt} {.65ex}
        \textbf{Repara} & \textbf{Method} & \textbf{Strategy} & \textbf{Backprop} & \textbf{Inference Overhead} \\
        \midrule
        \multirowcell{5}{LoRA and \\ Its Variants} 
        & LoRA & \makebox[25ex][r]{Rank-reducing\&-increasing} &  \ino & \ino \\
        & QLoRA & \makebox[25ex][r]{Freeze 4-bit quantized FM} &\ino &\ino \\
        & LoRA-FA & \makebox[25ex][r]{Freeze $\mathbb{A}'s$ weight}  &\ino &\ino \\
        & Incre-LoRA & \makebox[25ex][r]{Dynamically add parameters}  &\ino &\ino \\
        & Delta-LoRA & \makebox[25ex][r]{Add $\mathbb{A}$\&$\mathbb{B}'s$ delta} & \ino&\ino 
        \\
        \cmidrule(lr){1-5}
        MPO & decomposition & \makebox[25ex][r]{Train auxiliary tensors} &\ino &\ino  \\
        \specialrule{.16em}{.4ex}{0pt}
    \end{tabular}
    \end{adjustbox}
    \caption{Key comparison between parameterization PEFT. Backprop denotes whether reducing backpropagation costs. For Inference Overhead, $\ino$ means there is no extra overhead.}
    \label{tab: comparison_reparameterization}
\end{table}

\subsection{\textcolor{integrated}{Hybrid} PEFT}
A unique and promising approach in the PEFT field revolves around the integration of multiple methodologies. This strategic combination brings together several unique PEFT techniques such as LoRA~\cite{zhang2023lora}, BitFit~\cite{BenZaken2021BitFitSP}, P-Tuning~\cite{Liu2021GPTUT}, and others, into a singular strategic framework.
This integrative approach allows the model to draw on the strengths and insights of each methodology, thus establishing a comprehensive and robust framework. With this fusion, the model is primed to optimize parameters more efficiently, reduce computational burdens, and potentially enhance performance, providing an interesting and promising avenue in PEFT, as shown in Table~\ref{tab: comparison_integrated_methods}.

\sssection{E.1 Hybrid PEFT in Basics}

The main hybrid technique includes UniPELT~\cite{Mao2021UniPELTAU}, COMPACTER~\cite{Davison2021CompacterEL}, S$^4$~\cite{Chen2023ParameterEfficientFD}, NOAH~\cite{zhang2022neural}, and DiffFit~\cite{xie2023difffit}.

\textbf{UniPELT}~\cite{Mao2021UniPELTAU} is a unified framework integrating the core aspects of adapter~\cite{Houlsby2019ParameterEfficientTL}, prefix tuning~\cite{Li2021PrefixTuningOC}, and LoRA~\cite{zhang2023lora} and employing a gating mechanism to regulate these modules. The linear layer gating mechanism essentially decides each module's contribution and operation. The experimental results reveal that UniPELT consistently shows performance improvements spanning between 1\% and 4\% compared to the standalone PELT methods it integrates. In general, UniPELT supports the promise that integrated methods have in furthering the efficiency and effectiveness of allowing FMs to adapt to specific tasks.
\textbf{COMPACTER}~\cite{Davison2021CompacterEL} extended the concept of basic adapters by innovating on placement and training approaches, introducing a novel, lightweight adapter structure based on the Kronecker product of low-rank matrices. This advancement required an addition of merely 0.05\% to 0.2\% of the original model's parameters but yielded impressive performance on benchmarks such as GLUE~\cite{Wang2018GLUEAM} and SuperGLUE~\cite{Sarlin2019SuperGlueLF}.
In a basic adapter layer, represented by $k$ for hidden state size and $b$ for bottleneck size, two matrices of size ${k \times b}$ are typically involved. COMPACTER introduced parameterized hypercomplex multiplication layers, expressing each adapter's parameters as a Kronecker product of an ${n \times n}$ matrix $\mathbf{A}$ and a ${(k/n) \times (d/n)}$ matrix $\mathbf{B}$, significantly reducing the parameter count. Moreover, all adapters shared the matrix $\mathbf{B}$, which was decomposed into $n$ sets of two low-rank matrices, each sized ${(k/n) \times r}$ and ${r \times (d/n)}$, with $r$ set to 1 to minimize parameter count. Note that COMPACTER surpassed full fine-tuning when the training dataset was relatively small (0.1k - 4k instances).

\textbf{MAM adapter}~\cite{He2021TowardsAU} conducts a thorough investigation focusing on the arrangement of adapters and the employment of soft prompts corresponding to the quest to present a unified perspective on parameter-efficient transfer learning. They arrive at several implications and the key takeaways from their study include: (1) The scaled parallel adapter emerges as a standout candidate in modifying the FFN. (2) Adapters placed in parallel distinctly outperform those arranged sequentially. Moreover, directly comparing multi-head attention and FFN parallel placements demonstrates superior results. 
(3) In the context of restrained parameter budgets, modifications to the attention heads lead to optimal outcomes. Contrarily, FFN benefits the most from alterations when larger capacity settings are allowed. 
(4) Implementation of soft prompts, such as prefix tuning, yields significant performance advancements with tweaks to a minuscule 0.1\% of the parameters. 
Building on these insights, the MAM adapter introduces the multi-head attention adapter, a model that represents the integration of parallel adapters at the FFN layer and soft prompts. The model combines prefix modifications (with smaller bottleneck dimensions of $l=30$) implemented in attention sub-layers, and the scaled parallel adapter (with a bottleneck dimension of $r=512$) employed for modifying the FFN representation.
The MAM adapter exhibits a unique blend of efficiency and performance despite using only 6.7\% of the parameter count compared with full fine-tuning. Furthermore, it pulls ahead significantly compared to methods like BitFit and prompt tuning, consistently surpassing core methods such as LoRA, adapter, and prefix tuning.

${\textbf{S}^\textbf{4}}$~\cite{Chen2023ParameterEfficientFD} explored various ways to fine-tune models with fewer parameters. It looked at dividing layers into four groups, adjusting trainable parameters, selecting groups to fine-tune, and applying specific techniques. It introduced an innovative approach named $S^4$ that grouped layers into G1, G2, G3, and G4, resembling a spindle shape. The middle groups had more layers, while the top and bottom had fewer. All groups remained trainable with parameters spread uniformly across layers and different PEFT techniques were applied.
G1 used adapters, G2 benefited from adapters and prefix tuning, G3 was fine-tuned with adapters, prefix tuning, and BitFit, and G4 underwent prefix tuning, BitFit, and LoRA. Experiments show that the $S^4$ method with only 0.5\% of parameters consistently outperforms individual techniques in different models, sizes, and tasks.

\sssection{E.2 Hybrid PEFT in More FMs}

NOAH (Neural prOmpt seArcH)~\cite{zhang2022neural} implements neural architecture searches for designing prompt modules and incorporates adapter, LoRA, and VPT into each Transformer block.
DiffFit~\cite{xie2023difffit} only finetunes the bias terms and introduces scaling factors to achieve training efficiency and storage reduction. 
V-PEFT~\cite{yu2022towards} presents a unified analysis for PEFT approaches based on video tasks by investigating the fine-tuning position.
DreamBooth~\cite{ruiz2023dreambooth} utilizes a small number of images of an individual and introduces a new autogenous class-specific prior preservation loss to associate a distinct identifier with the subject while maintaining class variation.

\begin{table}[t!]
    \centering
    \renewcommand{\arraystretch}{1.0}
    \begin{adjustbox}{max width=\linewidth}
    \begin{tabular}{c|c|c|c}
        \specialrule{.16em}{0pt} {.65ex}
        \textbf{Hybrid Methods} & \textbf{Distinction} & \textbf{Backprop} & \textbf{Inference Overhead} \\
        \midrule
         \multirow{2}{*}{UniPELT} & \makebox[22ex][r]{\multirowcell{2}{LoRA\&Prefix Tuning\\\&Adapter}} & \multirow{2}{*}\ino & \multirow{2}{*}{FFN\&input} \\ 
         & & & \\
         \cmidrule(lr){1-4}
         COMPACTER & \makebox[22ex][c]{Adapters} &\ino & FFN \\ \cmidrule(lr){1-4}
         MAM Adapter & \makebox[22ex][r]{Adapters\&Soft Prompts} & \ino & FFN\&input \\ \cmidrule(lr){1-4}
         \multirow{2}{*}{$S^4$} & \makebox[22ex][r]{\multirowcell{2}{Adapters\&Prefix Tuning\\\&BitFit\&LoRA}} & \multirow{2}{*}\ino & \multirow{2}{*}{FFN\&input} \\ 
         & & & \\
        \specialrule{.16em}{.4ex}{0pt}
    \end{tabular}
    \end{adjustbox}
    \caption{Key comparison between hybrid PEFT. Backprop denotes whether reducing backpropagation costs. For Inference Overhead, FFN means adding overhead to the FFN part, and others are similar.}
    \label{tab: comparison_integrated_methods}
\end{table}

\sssection{E.3 Pros and Cons}

Here, we analyze the pros and cons of hybrid PEFT. This category of PEFT provides a unified framework to integrate various PEFT methods into a single and coordinated structure, thus enhancing the flexibility and adaptability of the overall system. Additionally, the hybrid approach can leverage the strengths of individual PEFT methods, leading to improved performance and robustness in handling diverse scenarios.
Nevertheless, some shortcomings also should be noted.

{\bf \(\bullet\) High complexity.} This category of PEFT may introduce heightened complexity, leading to increased computational demands, development costs, and labeling expenses. For instance, approaches like NOAH necessitate extra training for the super network and demand additional labeling efforts.

{\bf \(\bullet\) Limited performance.} The overall performance of this category might be constrained by unforeseen combinations, potentially resulting in suboptimal hybrid outcomes. Adjusting hyperparameters and managing the intricate interactions of different methods can present challenges,  that hinder the seamless integration of hybrid approaches.

\section{PEFT for Large Language Models}
\label{sec: application_llm}
\subsection{PEFT for Causal Language Models}

Causal LLMs are in vogue in the LLM community as a type of foundation language model~\cite{zhao2023survey}, also referred to as autoregressive LLM, e.g., GPT-3~\cite{floridi2020gpt}, BLOOM~\cite{le2023bloom}, Falcon~\cite{almazrouei2023falcon}, and LLaMA families~\cite{Touvron2023LLaMAOA}. Here, we briefly review the advances of the PEFT in causal LLMs. For instance, LLaMA-adapter~\cite{zhang2023llama} injects a set of learnable adaptation prompts after the transformer layers into the frozen LLaMA-7B~\cite{Touvron2023LLaMAOA}, which requires only 1.2M trainable parameters to extend the language instructions. Similarly, serial adapter tuning~\cite{Houlsby2019ParameterEfficientTL} and parallel adapter tuning~\cite{He2021TowardsAU} efficiently finetune GPT-J-6B~\cite{wang2022gpt} and BLOOM-7.1B~\cite{le2023bloom}, and outperform GPT-3.5 in mathematical reasoning. Additionally, LoRA families are often employed in this group of LLMs, e.g., QLoRA~\cite{Dettmers2023QLoRAEF} introduces a series of memory-saving parties to fine-tune LLaMA on Flan v2~\cite{longpre2023flan}, Alpace~\cite{alpaca} without sacrificing performance. LoRA-Sparse~\cite{song2024low} reduces more than half of the self-attention computations while enhancing NLP task performance based on LLaMA~\cite{Touvron2023LLaMAOA}. MoSLoRA~\cite{wu2024mixture} fuses MoE and LoRA to fine-tune LLaMA, improving commonsense reasoning. Moreover, Prefix tuning~\cite{Li2021PrefixTuningOC}, P-Tuning~\cite{Liu2021GPTUT}, and Prompt tuning~\cite{Lester2021ThePO} also support various causal LLMs, please refer to the open-source library for details.

\subsection{PEFT for Prefix Language Models}

Prefix LLMs, also known as non-causal LLMs~\cite{zhao2023survey}, are another mainstream in the LLM community, primarily represented by ChatGLM families~\cite{glm2024chatglm}. To recap, P-tuning series~\cite{Liu2021PTuningVP, Liu2021GPTUT} utilizes prompt tokens to fine-tune ChatGLM~\cite{glm2024chatglm} with only 0.1–0.3\% of trainable parameters as a generalized solution across various model scales and language understanding tasks. OrchMoE~\cite{wang2024orchmoe} utilizes a multiple-adapter modular skill architecture to fine-tune ChatGLM thus advancing the forward transfer during PEFT. At the same time, FATE-LLM~\cite{fan2023fate} leverages LoRA and P-Tuning v2 to tune ChatGLM-6B to evaluate language capabilities in a federated scenario, requiring just 0.06\% and 0.048\% trainable parameters, respectively. Similar efforts in this scenario include DP-LoRA~\cite{liu2023differentially}. While CPMI-ChatGLM~\cite{liu2024cpmi} applies P-Tuning v2 and LoRA to fine-tune ChatGLM-6B for a better understanding of real-world scenarios. MoELoRA~\cite{liu2023moelora} efficiently fine-tunes ChatGLM-6B by using task-driven gate functions to control the contribution of each LoRA.

Overall, we recapped the advances of PEFT methods in two representative types of foundational language models~\cite{zhao2023survey}: causal LLMs and prefix LLMs. In practice, encoder-decoder LLMs like T5~\cite{raffel2020exploring} are also one of the popular ones, the majority of the PEFT methods discussed above are equally available to them. For example, LLaMAFactory~\cite{zheng2024llamafactory} flexibly customizes a variety of PEFT schemes to enhance language modeling, such as LoRA~\cite{Hu2021LoRALA}, DoRA~\cite{liu2024dora}, rsLoRA~\cite{kalajdzievski2023rank}, PiSSA~\cite{meng2024pissa}, etc. The repository also covers multiple types of LLMs, including but not limited to the two types we discussed.

\section{PEFT for Visual Foundation Models}
\label{sec: application_vis_found}
\subsection{PEFT for Basic Vision Models}

ViT is the prevailing and basic backbone of VFMs. Accordingly, this subsection attention to recent advances of PEFT in ViT. In a broad sense, the VFM in this category only takes into account images as inputs. Specifically, a range of PEFT approaches have been considered for VFMs, such as adapter tuning (AdaptFormer~\cite{chen2022adaptformer}, Convpass~\cite{jie2022convolutional}, AIM~\cite{yang2023aim}, ST-Adapter~\cite{pan2022st}, Rob-Adapter~\cite{sharma2023lossless}, LoRand~\cite{yin20231}, SCT~\cite{zhao2024sct}, Polyhistor~\cite{liu2022polyhistor}, VMT-Adapter~\cite{xin2024vmt}), prompt tuning (VPT~\cite{jia2022visual}, CVP~\cite{tsai2023convolutional}, LPT~\cite{dong2023lpt}, IDPT~\cite{zha2023instance}, Pro-tuning~\cite{nie2023pro}, LION~\cite{wang2024lion}, ViPT~\cite{zhu2023visual}, VP~\cite{bahng2022exploring}, EVP~\cite{wu2022unleashing}, DAM-VP~\cite{huang2023diversity}, EVP-L~\cite{liu2023explicit}, ProSFDA~\cite{hu2022prosfda}, P2P~\cite{wang2022p2p}, ILM-VP~\cite{chen2023understanding}), prefix tuning (Prefix-tuning~\cite{Li2021PrefixTuningOC}, PATT~\cite{yu2022towards}, eTT~\cite{xu2023exploring}, LAM~\cite{gao2023unified}, VQT~\cite{tu2023visual}), side tuning (Side-Tuning~\cite{zhang2020side}, SAN~\cite{xu2023side}, ViT-Adapter~\cite{chen2022vision}, LST~\cite{Sung2022LSTLS}, SAM-LST~\cite{chai2023ladder}, E$^{3}$VA~\cite{yin2023parameter}, CSN (DTL)~\cite{fu2024dtl}), specification tuning (Linear Probe~\cite{radford2021learning}, BitFit~\cite{BenZaken2021BitFitSP}, DP-BiTFiT~\cite{bu2022differentially}, DiffFit~\cite{xie2023difffit}, LN-TUNE~\cite{basu2024strong}), and reparameter tuning (LoRA~\cite{Hu2021LoRALA}, KAdaptation~\cite{he2023parameter}, FacT~\cite{jie2023fact}, EFFT~\cite{chen2023aggregate}, SSF~\cite{lian2022scaling}, RepAdapter~\cite{luo2023towards}, ATTNSCALE~\cite{basu2024strong}, PHNNs~\cite{grassucci2022phnns}, DnA~\cite{jiang2022dna}), etc.

As mentioned above, various PEFT methods are widely presented in the downstream tasks of VFMs. Like, i) image recognition is the primary scenario for PEFT, such as AdaptFormer~\cite{chen2022adaptformer}, VPT~\cite{jia2022visual}, CSN (DTL)~\cite{fu2024dtl}. Rob-Adapter~\cite{sharma2023lossless} proposes lossless adaptation for achieving optimal performance in manipulation tasks. Moreover, quite a few works have also made successful efforts for image-related scenarios, like LPT~\cite{dong2023lpt}, FacT~\cite{jie2023fact}, LoRA~\cite{Hu2021LoRALA}, NOAH~\cite{zhang2022neural}, MONA~\cite{yin20245}, etc. ii) PEFT is also influential in video understanding. Among them, AdaptFormer~\cite{chen2022adaptformer}, VPT~\cite{jia2022visual}, and LoRA~\cite{Hu2021LoRALA} have been big sellers in video-related tasks. ST-adapter~\cite{pan2022st} only requires a small ($\sim$8\%) per-task parameter cost to understand the video. AIM~\cite{yang2023aim} proposes spatial-, temporal- and joint-adaptation with substantially fewer tunable parameters for efficient video understanding. APT~\cite{bandara2024attention} involves attention prompt tuning with less than 1\% of the parameters to reduce latency and FLOPs in video recognition. Moreover, LoSA~\cite{gupta2024losa}, RaSTFormer~\cite{zhang2024rastformer}, etc. also resorted to efforts in temporal action localization and short videos.

\subsection{PEFT for Prompted Vision-Language Models}

This subsection attention to recent advances in PEFT for the prompted VLMs. In general, the VFM in this category takes into account visual and textual information as inputs. In details, a series of PEFT approaches have been applied for the prompted VLMs, such as visual grounding (CoOp~\cite{zhou2022learning}, CoCoOp~\cite{zhou2022conditional}, ProGrad~\cite{zhu2023prompt}, MaPLe~\cite{khattak2023maple}, TPT~\cite{shu2022test}, CPT~\cite{yao2024cpt}, DiffTPT~\cite{feng2023diverse}, CLIP-Adapter~\cite{gao2024clip}, Tip-Adapter~\cite{zhang2021tip}, PromptSRC~\cite{khattak2023self}, BadCLIP~\cite{bai2024badclip}, MePT~\cite{wang2024mept}, NODE-Adapter~\cite{zhang2024node}, AAPL~\cite{kim2024aapl}, CoPL~\cite{goswami2024copl}, Any-Shift Prompting~\cite{xiao2024any}, PIN~\cite{dorkenwald2024pin}, CLAP~\cite{silva2024closer}, TCP~\cite{yao2024tcp}, DePT~\cite{zhang2024dept}), semantic segmentation (SAN~\cite{xu2023side}, LLMFormer~\cite{shi2024llmformer}, FC-CLIP~\cite{yu2024convolutions}, MasQ-Tuning~\cite{xu2023masqclip}, Test Time Prompt Tuning (TTPT from FreeSeg)~\cite{qin2023freeseg}, mask prompt tuning~\cite{liang2023open}, EVP~\cite{liu2023explicit}, ETRIS~\cite{xu2023bridging}), video understanding (Vita-CLIP~\cite{wasim2023vita}, MA-CLIP~\cite{xing2023multimodal}, DualPath~\cite{park2023dual}, Text-Adapter (M$^{2}$-CLIP)~\cite{wang2024multimodal}, TDS-CLIP~\cite{wang2024tds}, OmniCLIP~\cite{liu2024omniclip}, EVL~\cite{lin2022frozen}, Side4Video~\cite{yao2023side4video}, EZ-CLIP~\cite{ahmad2023ez}, ActPrompt~\cite{wang2024actprompt}, MV-Adapter~\cite{jin2024mv}), point cloud segmentation (PointCLIP V2~\cite{zhu2023pointclip}, P2P~\cite{wang2022p2p}, CLIP2Point~\cite{huang2023clip2point}, EPCL~\cite{huang2024frozen}, IDPT~\cite{zha2023instance}, DAPT~\cite{zhou2024dynamic}), etc.

According to the type of prompts input to the model, the existing work is roughly divided into textual and visual prompt VLM. \textit{i) textual prompt:} a series of works (e.g., CoOp~\cite{zhou2022learning}, KgCoOp~\cite{yao2023visual}) uses prompt tuning methods for textual inputs to perform PEFT in vision tasks. 
TCP~\cite{yao2024tcp} uses textual-based class-aware prompts to unlock limited generalizations of textual tokens to unseen domains. 
Note that some of the methods in this group are initially proposed for textually prompted VLM, though they are also commonly used in more generalized VLM. \textit{ii) visual prompt:} this category of PEFT methods (e.g., OVSeg~\cite{liang2023open} and CPT~\cite{yao2024cpt}) requires image and visual or textual prompts to perform fine-tuning~\cite{sun2024vrp}, and these generally include visual prompts (point, bounding box, mask, color), text prompts, reference prompts, composition, etc. 
GP-SAM and VRP-SAM~\cite{sun2024vrp}, etc. encode various visual references and geometric prompts (point, box, scribble, mask) into prompts embedding as inputs to segment anything. PIN~\cite{dorkenwald2024pin} presents a visual prompt method that an input-agnostic positional insert to explores the localization ability of visual grounding. In brief, this category of PEFT methods follows the principle that customizes different visual tasks and prompts.

\subsection{PEFT for Visual Content Generation Models}

\begin{figure}[t!] 
\centering 
\includegraphics[width=0.49\textwidth]{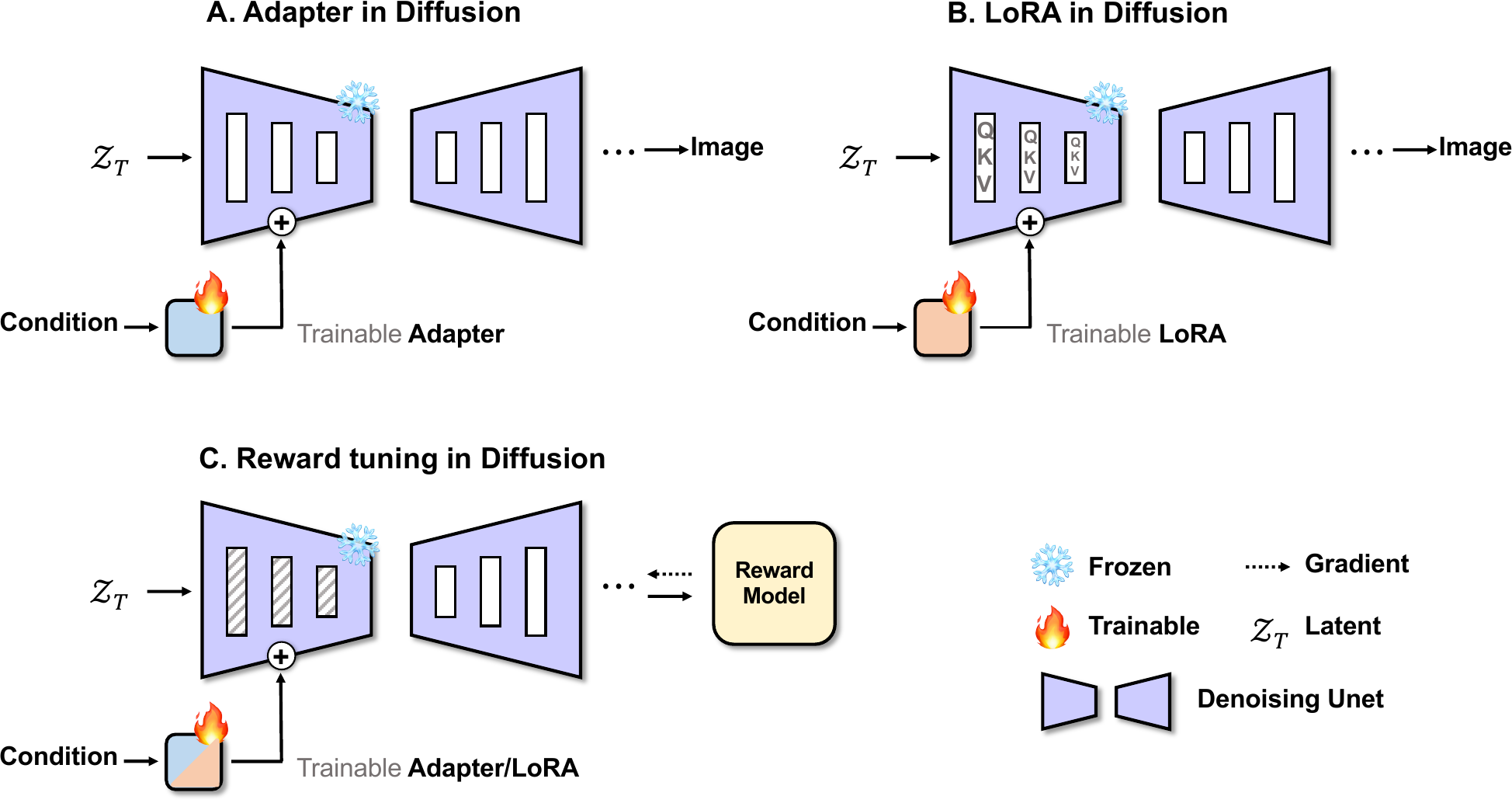} 
\captionsetup{justification=raggedright, singlelinecheck=false}
\caption{Prevailing PEFT in VGMs. A. Adapter tuning in diffusion model; B. LoRA tuning in diffusion model; C. Reward tuning in diffusion model.} 
\label{fig: figure_SD} 
\end{figure}

Recently, the diffusion model has been trending as FMs in visual content generation. In this subsection, we review recent advances in PEFT methods for diffusion models, as shown in Fig.~\ref{fig: figure_SD}. Specifically, a range of PEFT methods are implemented in various diffusion model scenarios. For instance, image generation (Textual Inversion~\cite{gal2022image}, T2I-Adapter~\cite{mou2024t2i}, DreamBooth~\cite{ruiz2023dreambooth}, ControlNet~\cite{zhang2023adding}, GLIGEN~\cite{li2023gligen}, Uni-ControlNet~\cite{zhao2024uni}, ControlNeXt~\cite{peng2024controlnext}, CCM~\cite{xiaoccm}, IP-Adapter~\cite{ye2023ip}, CTRL-Adapter~\cite{lin2024ctrl}, X-Adapter~\cite{ran2024x}, LoRA-Composer~\cite{yang2024lora}, DiffuseKronA~\cite{marjit2024diffusekrona}, SVDiff~\cite{han2023svdiff}, SODA~\cite{hudson2024soda}), video generation (SimDA~\cite{xing2024simda}, StyleCrafter~\cite{liu2023stylecrafter}, I2V-Adapter~\cite{guo2024i2v}, Still-Moving~\cite{chefer2024still}, Tune-A-Video~\cite{wu2023tune}, CTRL-Adapter~\cite{lin2024ctrl}, Customize-A-Video~\cite{ren2024customize}, ControlNeXt~\cite{peng2024controlnext}), editing (Concept Sliders~\cite{gandikota2023concept}, PTI~\cite{dong2023prompt}, CCEdit~\cite{feng2024ccedit}, SVDiff~\cite{han2023svdiff}, DiffMorpher~\cite{zhang2024diffmorpher}), super-resolution (ResAdapter~\cite{cheng2024resadapter}, DiffFit~\cite{xie2023difffit}, ControlNeXt~\cite{peng2024controlnext}), 3D generation (IPDreamer~\cite{zeng2023ipdreamer}), etc. Among these methods, LoRA~\cite{Hu2021LoRALA}, ControlNet~\cite{zhang2023adding}, and Adapter related approaches~\cite{mou2024t2i, guo2024i2v, cheng2024resadapter} are frequently employed in various diffusion models. Whereas the trends of PEFT in various scenarios are analyzed, image generation and video generation are obviously preferred more.

In detail, ControlNet series~\cite{zhang2023adding} tuning trainable copies to learn various controllable conditions, e.g., Openpose, Depth, Canny, Lineart, AnimeLineart, Mlsd, Scribble, Hed, Pidi, Teed, Segment, Norma, and their permutations. LoRA-related techniques~\cite{Hu2021LoRALA, yang2024lora} are noted in image or video generation, editing, etc. such as Smooth Diffusion~\cite{guo2024smooth}, STAMIINA, DreamSync~\cite{sun2023dreamsync}, StyleAdapter~\cite{wang2023styleadapter}, Mix-of-Show~\cite{gu2024mix}, and DragVideo~\cite{deng2023dragvideo}. Broadly speaking, LoRA~\cite{Hu2021LoRALA} is normally configured in the attention module whereas more effort is given to the temporal cross-frame attention in stable video diffusion, like T-LoRA~\cite{ren2024customize} from Customize-A-Video. Adapter-related techniques~\cite{mou2024t2i, guo2024i2v, cheng2024resadapter} prefer to introduce a variety of single or combined lightweight adapter modules to fine-tune the diffusion model for precise control of various conditions.

\section{PEFT for multi-modal foundation models}
\label{sec: application_mm}
\subsection{PEFT for Broadly Multi-Modal Foundation Models}

In a narrow sense, some VLMs mentioned in the former subsection subsume the scope of the multi-modal model as they involve both text and vision. Nevertheless, the above models emphasize more on individual skills of visual tasks, e.g., grounding, and segmentation. Thus, we review them in the scope of vision. Here, we survey the PEFT approach in the broadly MFMs, which is not bound to a single language or visual skill but to a broader multi-modal understanding. For example, PEFT-MLLMs~\cite{zhou2024empirical} execute an empirical explore on Adapter, LoRA, Prefix tuning, IA3 for LLaVA-1.5, ShareGPT4V, Qwen-VL-Chat. LLaMA-Adapter V2~\cite{gao2023llama} unlocks more learnable parameters to enhance efficiently LLaMA-Adapter~\cite{zhang2023llama}, thus, open-ended multi-modal instructions are executed by merely inserting the 14M parameter (0.04\%) on LLaMA. LayerNorm Tuning~\cite{zhao2023tuning} only tuning LayerNorm within each attention block is ample to improve the multimodal performance. 
LoRA-Sparse~\cite{song2024low} introduces low-rank linear projection layers for sparse attention to boost the multi-modal performance of LLaVA-1.5~\cite{cai2024vip}.
Moreover, LoRA~\cite{Hu2021LoRALA} and Q-Former~\cite{li2023blip} prevail in Monkey~\cite{li2024monkey}, mPLUG-Owl~\cite{ye2023mplug}, CogVLM~\cite{wang2023cogvlm, hong2024cogvlm2}, and GLM-4V~\cite{hong2024cogvlm2}, etc. to enhance different multimodal capabilities.

\subsection{PEFT for NExT Multi-Modal Foundation Models}

Next-generation MFMs~\cite{wu2023next} are not limited to a few modalities, they can perceive inputs and generate outputs in any combo of text, image, video, and audio, like CoDi series~\cite{tang2024any, tang2024codi}, HuggingGPT~\cite{shen2024hugginggpt}, Visual-ChatGPT~\cite{wu2023visual}, SEED-X~\cite{ge2024seed}, Gemini 1.5 Pro~\cite{reid2024gemini}, Show-o~\cite{xie2024show}, and NExT-GPT~\cite{wu2023next}. Here, we investigate the recent advances of PEFT in this category. For example, SEED-X~\cite{ge2024seed} pre-trained first on Llama2-chat-13B and then used LoRA~\cite{Hu2021LoRALA} on massive multi-modal data. Anole~\cite{chern2024anole} utilizes a data-efficient (about 6000 samples) and parameter-efficient (fewer than 40M parameters) fine-tuning strategy to facilitate visual and multimodal generation. NExT-GPT~\cite{wu2023next} likewise employs LoRA~\cite{Hu2021LoRALA} to adjust a fairly few parameters (1\%) to update a specific projection layer thus enhancing the multi-modal capability.
\section{Discussion and Future Directions}
\label{sec: discussion}

\subsection{Observation of Current Trend}

\textbf{Reliability.} PEFT methods are sensitive to hyperparameters, e.g. bottleneck dimensions, rank, and layer order. Furthermore, due to the structures or networks used in PEFT, which are significantly smaller than the FM itself, optimal hyperparameters often differ substantially from those used for full fine-tuning. For example, the optimal learning rate for PEFT is typically much higher than that for full fine-tuning. Therefore, developing simple and efficient hyperparameter solutions with low sensitivity is crucial.

\textbf{Interpretability.} Understanding the internal mechanisms of PEFT methods remains a challenge. In LLMs, prompts can be explained in a relatively intuitive way. However, in FMs, a primary challenge is that various prompts are learned as unordered token-based prompts, which are difficult to translate into an understandable format. Additionally, different PEFT methods confront specific interpretability challenges. For example, understanding the relationship between learned parameters and layers in adapters is an important topic.

\textbf{Unified Benchmark.} Despite the availability of libraries like Hugging Face's PEFT and AdapterHub, there remains a notable lack of comprehensive benchmarks for PEFT. Different studies use varied evaluation datasets and task setups, leading to inconsistent performance assessment standards, which in turn impacts user to evaluate the strengths and weaknesses of different PEFT methods. To tackle this, a current trend is to establish standardized baselines to facilitate fairer comparisons across methods.

\vspace{-0.2cm}

\subsection{Future Directions}

\textbf{Across Disciplines.} The future advancement in PEFT is likely to arise from interdisciplinary insights, especially as FMs are applied to fields ranging from medicine and natural sciences to social sciences. 
In particular, integrating domain-specific constraints into the PEFT framework may lead to more tailored fine-tuning approaches. 
For example, in medical imaging, incorporating medical domain knowledge and low-dimensional priors or causal relationships could enhance model performance, even with minimal parameter updates.

\textbf{Continual PEFT.} PEFT provides a well-performed solution for fine-tuning FMs on specific tasks. Nevertheless, when these methods are adapted to a sequence of tasks or dynamic data flow, the model may interfere with or overwrite learned knowledge. In contrast, continual learning focuses on developing systems that can continuously learn new tasks while retaining memory and performance on learned tasks. The combination of PEFT and continual learning would make PEFT more robust in dynamically changing tasks or environments. Thus, developing PEFT for CL may potentially contribute to smarter learning systems in the real world.

\textbf{Architecture for PEFT.} Understand the applicability and advantages of specific architectures for PEFT and explore how to design more effective PEFT schemes for particular architectures. For example, analyze the response characteristics of different layers and components in the Transformer architecture to PEFT, providing a basis for architecture optimization and customized PEFT methods.

\textbf{Scaling Laws of PEFT.} Current efforts reveal diminishing returns beyond a certain threshold of trainable parameters, indicating an optimal range for parameter selection. For PEFT methods, understanding these scaling behaviors is crucial to optimizing efficiency and guiding future research. For example, how does performance scale when increasing or decreasing the number of trainable parameters in PEFT methods such as LoRA, adapters, or prefix-tuning?  This can provide guidance for future model design and fine-tuning strategies.

\textbf{Layered Abstraction.} The layered abstraction in PEFT parallels how the human brain processes and stores information hierarchically. In the brain, sensory inputs are processed through layers of increasing complexity, from lower-level sensory neurons up to higher-order cognitive regions. This layered approach enables the brain to create abstract representations and make sense of complex information. Similarly, PEFT often works by adjusting parameters at different levels of the model, such as early layers for general features and later layers for task-specific adaptation. By fine-tuning specific layers or adding modular structures, PEFT facilitates a nuanced, hierarchical adaptation to tasks—mirroring the brain's ability to build from simple to complex representations. This layered design not only improves model flexibility but also allows for efficient reuse of existing knowledge across tasks.

\textbf{Brain-Inspired PEFT.} Interestingly, PEFT aligns with principles in neuroscience, particularly with theories of efficient coding and synaptic plasticity. In the brain, adaptation and learning occur through mechanisms that prioritize energy efficiency while maintaining flexibility and robustness—a concept that resonates with the goals of PEFT. For example, in the human brain, when we learn something new, rather than adjusting all neural connections, only specific synaptic pathways are modified. This selective adjustment helps efficiently incorporate new information without drastically disrupting existing knowledge. Similarly, PEFT allows models to specialize and adapt to new tasks by updating a minimal number of parameters, aligning with how neural circuits in the brain reorganize for new skills or experiences. This resemblance opens intriguing opportunities for incorporating biologically inspired mechanisms, which could lead to more biologically plausible and efficient fine-tuning processes.

\section{Conclusion}
\label{sec: conclusion}
In conclusion, the integration of PEFT with FMs showcases a promising avenue for efficient model adaptation across various tasks and domains. As highlighted in this survey, the rapid evolution of FMs and the active PEFT community underscore the importance of staying abreast of technological trends for optimal performance. By exploring adaptation strategies such as \textcolor{selective}{Selective}, \textcolor{additive}{Additive}, \textcolor{prompt}{Prompt}, \textcolor{reparameter}{Reparameterization}, and \textcolor{integrated}{Hybrid} PEFT, and across different model structures (e.g., LLM, VFM, VLM, MFM, and VGM), this survey offers insights into enhancing efficiency and effectiveness. The survey emphasizes the need for a systematic understanding of PEFT techniques in the context of diverse FMs, paving the way for future advancements and applications in the field.

\bibliographystyle{IEEEtran}
\bibliography{references}

\end{document}